\documentclass{article}
\usepackage{times}
\usepackage{graphicx} 
\usepackage{subfigure} 
\usepackage{footnote}
\usepackage{amsfonts}
\usepackage{mdframed}

\usepackage{natbib}
\usepackage{comment}
\usepackage{multirow}

\usepackage[small,tight]{bibhacks}

\newlength\llength
\usepackage{iclr2015,times}
\iclrconference

\usepackage{hyperref}
\usepackage{url}

\title{Learning to Execute}

\usepackage{float}

\floatstyle{plain} 
\newfloat{code}{H}{myc}

\usepackage[]{algorithm2e}
\usepackage[]{algorithmic}
\usepackage{amsmath}
\usepackage{graphics}
\usepackage{epsfig}
\usepackage{amsmath}
\usepackage{listings}

\usepackage{mathtools}

\usepackage{paralist}

\makeatletter
\ifcase \@ptsize \relax
  \newcommand{\miniscule}{\@setfontsize\miniscule{4}{5}}
\or
  \newcommand{\miniscule}{\@setfontsize\miniscule{5}{6}}
\or
  \newcommand{\miniscule}{\@setfontsize\miniscule{5}{6}}
\fi
\makeatother

\newcommand {\aplt} {\ {\raise-.5ex\hbox{$\buildrel<\over\sim$}}\ }

\newcommand{\BigO}[1]{\ensuremath{\operatorname{O}\left(#1\right)}}
\usepackage[symbol*]{footmisc}

\DefineFNsymbolsTM{myfnsymbols}{
  \textasteriskcentered *
  \textdagger    \dagger
  \textdaggerdbl \ddagger
  \textsection   \mathsection
  \textbardbl    \|%
  \textparagraph \mathparagraph
}%

\usepackage{hyperref}

\begin{document} 

\author{
Wojciech Zaremba\footnote{}\footnotetext{Work done while the author was in Google Brain.} \\
New York University\\
\texttt{woj.zaremba@gmail.com} \\
\And
Ilya Sutskever \\
Google \\
\texttt{ilyasu@google.com} \\
}

\maketitle

\begin{abstract}

Recurrent Neural Networks (RNNs) with Long Short-Term Memory units (LSTM) are widely used because
they are expressive and are easy to train.   Our interest lies in
empirically evaluating the expressiveness and the learnability of LSTMs in the sequence-to-sequence regime by training 
them to evaluate short computer programs, a domain that has traditionally been seen as too complex
for neural networks. We consider a simple class of programs 
that can be evaluated with a single left-to-right pass using constant memory.
Our main result is that LSTMs can learn to map the character-level representations of such programs to their correct outputs. 
Notably, it was necessary to use curriculum learning,  and while conventional curriculum learning proved ineffective, we
developed a new variant of curriculum learning that improved our networks' performance in all experimental conditions.
The improved curriculum had a dramatic impact on an addition problem, making it possible to train an LSTM to add two 9-digit
numbers with 99\% accuracy.

\end{abstract} 

{\textcolor{white} {\fontsize{0.001cm}{0.01em}\selectfont \footnotemark\footnotetext{Work done while the author was in Google Brain.} }}

\section{Introduction}
\vspace{-2mm}

Execution of computer programs requires dealing with a number of
nontrivial concepts.  To execute a program, a system has to understand
numerical operations, if-statements, 
variable assignments, the compositionality of operations, and many more.

We show that Recurrent Neural Networks (RNN) with Long Short-Term Memory (LSTM) units
can accurately evaluate short simple programs in the sequence-to-sequence framework
of \cite{sutskever2014sequence}.
The LSTM reads the program
character-by-character and computes the program's output. 
We consider a constrained set of computer programs that can be evaluated in 
linear time and constant memory, because the LSTM reads the program only once and its memory
capacity is limited (Section \ref{sec:ffcm}). 

We found it difficult to train LSTMs to execute computer programs, 
so we used curriculum learning to simplify the learning problem. We design a curriculum procedure which
outperforms both conventional training that uses no curriculum learning (\textit{baseline}) as well as  
the \textit{naive} curriculum learning of strategy of \cite{bengio2009curriculum} (Section \ref{sec:curriculum}).
We provide a plausible explanation for the effectiveness of our procedure relative to \textit{naive} curriculum
learning (Section \ref{sec:hypothesis}).

Finally, in addition to curriculum learning strategies, 
we examine two simple input transformations that further simplify the sequence-to-sequence learning problem. We show that,
in many cases, reversing the input sequence \citep{sutskever2014sequence} and replicating the input sequence improves the LSTM's
performance on a memorization task (Section \ref{sec:copy}).

The code for replicating most of the experiments in this work can be found in
\url{https://github.com/wojciechz/learning_to_execute}.

\section{Related work}
\vspace{-2mm}
There has been related research that used Tree Neural Networks (also known as Recursive Neural Networks)
to evaluate symbolic mathematical expressions
and logical formulas \citep{zaremba2014learning, bowman2014recursive, bowman2013can}, 
which is close in spirit to our work. 
Computer programs are more complex than mathematical or logical expressions 
because it is possible to simulate either with an appropriate computer program.

From a methodological perspective, we formulate the program evaluation task 
as a sequence-to-sequence learning problem with a recurrent neural network \citep{sutskever2014sequence}
(see also \citep{mikolov2012statistical, sutskever2013training, pascanu2013construct}). 
Other interesting applications of recurrent neural networks include  
speech recognition \citep{robinson1996use, graves2013speech},
machine translation \citep{cho2014learning, sutskever2014sequence}, 
handwriting recognition \citep{pham2013dropout,zaremba2014recurrent}, and many more.

\cite{maddison2014structured} trained a language model of program text,
and \cite{mou2014building} used a neural network to determine whether two programs are equivalent.
Both of these approaches require the parse trees of programs, while the input to our model is a string of character representing our program.

Predicting program output requires that the model deals with long term dependencies  
that arise from variable assignment. For this reason, we chose to use the Long Short-Term Memory model \citep{hochreiter1997long}, although there are many other 
RNN variants that perform well on tasks with long term dependencies 
\citep{cho2014learning, jaeger2007optimization, koutnik2014clockwork, martens2010deep, bengio2013advances}. 

Initially, we found it difficult to train LSTMs to accurately evaluate programs.
The compositional nature of computer programs suggests that the LSTM would learn
faster if we first taught it about the individual operators and 
how to combine them.  This approach can be implemented with curriculum learning
 \citep{bengio2009curriculum, kumar2010self, lee2011learning}, which prescribes
to gradually increase the ``difficulty level'' of the examples presented to the LSTM.
It is partially motivated by fact that humans and animals learn much faster when
they are given hard but manageable tasks. Unfortunately, we found the \textit{naive}
curriculum learning strategy of \citet{bengio2009curriculum} to sometimes be
harmful. One of our key contributions is the formulation of a new curriculum
learning strategy that substantially improves the speed and the quality of training in every 
experimental setting that we considered.

\begin{figure}[t!]
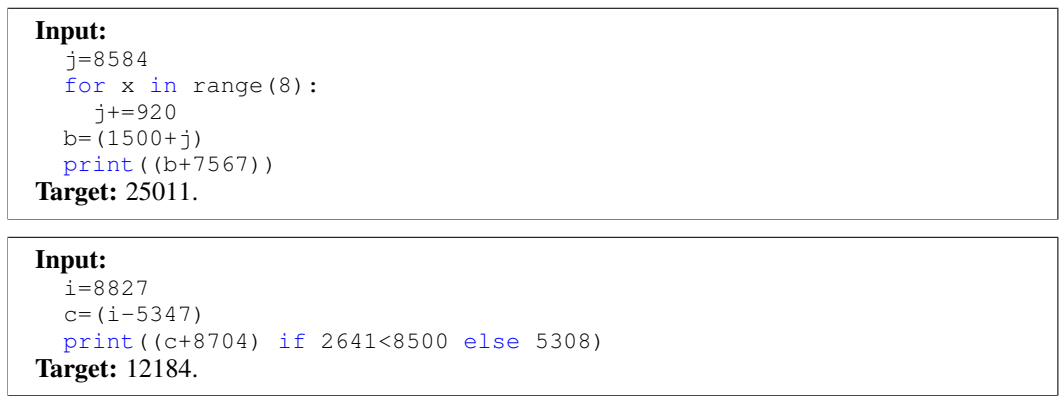

\tiny{
  \begin{code}
  \begin{mdframed}
  {\bf Input:}
  \begin{lstlisting}
  j=8584
  for x in range(8):
    j+=920
  b=(1500+j)
  print((b+7567))
  \end{lstlisting} 
  {\bf Target:} 25011. 
  \end{mdframed}
  \end{code}
\vspace{-10mm}
  \begin{code}
  \begin{mdframed}
  {\bf Input:}
  \begin{lstlisting}
  i=8827
  c=(i-5347)
  print((c+8704) if 2641<8500 else 5308)
  \end{lstlisting}
  {\bf Target:} 12184.
  \end{mdframed}
  \end{code}
\vspace{-5mm}
  \caption{Example programs on which we train the LSTM. The output of each program is a single integer. 
    A ``dot'' symbol indicates the end of the integer, which has to be predicted by the LSTM.}
  \label{fig:example}
}
\end{figure}

\section{Program Subclass}
\vspace{-2mm}
\label{sec:ffcm}

We train RNNs on the class of short programs that can be evaluated in $\BigO{n}$ time and constant memory. 
This restriction is dictated by the computational structure of the RNN itself, 
as it can only perform a single pass over the program and its memory is limited.
Our programs use the Python syntax and are constructed from a small number of operations and their compositions (nesting).
We allow the following operations: addition, subtraction, multiplication, variable assignments, if-statements, and
for-loops, but we forbid double loops. Every program ends with a single ``print'' statement whose output is an integer. 
Two example programs are shown in Figure \ref{fig:example}.

We select our programs from a family of distributions parametrized by their \textit{length} and \textit{nesting}. 
The \textit{length} parameter is the number 
of digits in the integers that appear in the programs (so the integers are chosen uniformly from $[1, 10^{\mathrm{length}}]$).    
The appendix presents the pseudocode \ref{alg:prog_gen} of the algorithm used to generate our programs.
For example, two programs that are generated with $\text{length}=4$ and $\text{nesting}=3$ are shown in Figure \ref{fig:example}.

We impose restrictions on the operands of multiplication and on the ranges of for-loop, since they pose a greater difficulty to our model.
We constrain one of the arguments of multiplication and the range of for-loops to be chosen uniformly from 
the much smaller range $[1, 4\cdot\text{ length}]$. 
We do so since our models are able to perform linear-time computation while generic 
integer multiplication requires superlinear time. Similar considerations apply to for-loops, 
since nested for-loops can implement integer multiplication. 

The \textit{nesting} parameter is the number of times we are allowed to combine the operations with each other. 
Higher values of \textit{nesting} yield programs with deeper parse trees. 
Nesting makes the task much harder for the LSTMs, because they do not
have a natural way of dealing with compositionality, unlike Tree Neural Networks. 
It is surprising that the LSTMs can handle nested expressions at all.
The programs also do not receive an external input.

It is important to emphasize that the LSTM reads the entire input one character at a time and produces the output one character at a time. The 
characters are initially meaningless from the model's perspective; for instance, the model does not know that ``+'' means addition or that $6$ is followed by $7$.
In fact, scrambling the input characters (e.g., replacing ``a''  with ``q'', ``b'' with ``w'', etc.,) 
has no effect on the model's ability to solve this problem.
We demonstrate the difficulty of the task by presenting an input-output example with scrambled characters in Figure \ref{fig:scr_example}. 

\begin{figure}[t!]
\footnotesize{
\begin{code}
\begin{mdframed}
{\bf Input:}
\begin{lstlisting}
vqppkn
sqdvfljmnc
y2vxdddsepnimcbvubkomhrpliibtwztbljipcc
\end{lstlisting}
{\bf Target:} hkhpg
\end{mdframed}
\end{code}
\vspace{-10mm}
\caption{A sample program  with its outputs when  the characters are scrambled.  It helps illustrate the difficulty faced by our neural network.   }
\label{fig:scr_example}
}
\end{figure}

Finally, we wanted to verify that our program are not trivial to evaluate, by ensuring that the bias coming from
Benford's law \citep{hill1995statistical} is not too strong. Our setup has $12$ possible output characters, that is
$10$ digits, the end of sequence character, and minus. Their output distribution is not uniform, 
which can be seen by noticing that the minus sign and the dot do not occur with the same frequency as
the other digits. If we assume that the output characters are independent, the probability
of guessing the correct character is $\sim 8.3\%$. The most common character is $1$ which occurs with probability
$12.7\%$ over the entire output.

However, there is a bias in the distribution of the first character. There are $11$ possible choices, 
which can be randomly guessed with a probability of $9\%$. The most common character is $1$, and it occurs with a probability
$20.3\%$ in its first position, indicating a strong bias. Still, this value is far below our model prediction accuracy.
Moreover, the most probable second character in the first position of the output occurs with probability $12.6\%$, which is indistinguishable from
probability distribution of digits in the other positions. The last character is always the end of sequence. The most common digit prior to the last character
is $4$, and it occures with probability $10.3\%$.  
These statistics are computed with $10000$ randomly generated programs with $length=4$ and $nesting=1$.
The absence of a strong bias for this configuration suggests that there will be even less bias in with greater nesting and longer digits,
which we have also confirmed numerically.

\subsection{Addition Task}
\vspace{-2mm}
\label{sec:add}

It is difficult to intuitively assess the accuracy of an LSTM on a program evaluation task. 
For example, it is not clear whether an accuracy of  $50\%$  
is impressive. Thus, we also evaluate our models on a more familiar addition task, where the difficulty is measured by the 
length of the inputs.  We consider the addition of only two numbers of the same length (Figure \ref{fig:addition})
that are chosen uniformly from $[1, 10^{length}]$. Adding two number of the same length is simpler than adding variable
length numbers. Model doesn't need to align them.

\begin{figure}
\begin{code}
\begin{mdframed}
{\bf Input:}
\begin{lstlisting}
print(398345+425098)
\end{lstlisting}
{\bf Target:} 823443
\end{mdframed}
\end{code}
\vspace{-10mm}
\caption{A typical data sample for the addition task.}
\label{fig:addition}
\end{figure}

\subsection{Memorization Task}
\vspace{-2mm}
\label{sec:copy}

In addition to program evaluation and addition, we also investigate the task of memorizing a random sequence of numbers. 
Given an example input $123456789$, the LSTM reads it one character at a time,
stores it in memory, and then outputs $123456789$ one character at a time. 
We present and explore two simple performance enhancing techniques:
input reversing \citet{sutskever2014sequence} and input doubling.

The idea of input reversing is to reverse the order of the input ($987654321$) while keeping the desired output unchanged
 ($123456789$). It may appear to be a neutral operation because the average distance between
each input and its corresponding target does not change. 
However, input reversing introduces many short term
dependencies that make it easier for the LSTM to learn to make correct predictions. 
This strategy was first introduced by \citet{sutskever2014sequence}.

The second performance enhancing technique is input doubling, where we present the 
input sequence twice  (so the example input becomes $123456789;123456789$), 
while the output remains unchanged ($123456789$). This method is meaningless from a probabilistic perspective
 as RNNs approximate the conditional distribution $p(y|x)$, yet here we attempt to learn $p(y|x, x)$. Still, it gives noticeable
performance improvements. By processing the input several times before producing the output, 
the LSTM is given the opportunity to correct any mistakes or omissions it made before.

\section{Curriculum Learning}
\vspace{-2mm}
\label{sec:curriculum}

Our program generation procedure is parametrized by \textit{length} and \textit{nesting}. 
These two parameters allow us control the complexity of the program. When 
 \textit{length} and \textit{nesting} are large enough, the learning problem becomes nearly intractable. 
This indicates that in order to learn to evaluate 
programs of a given $\text{\textit{length}}=a$ and $\text{\textit{nesting}}=b$,
it may help to first learn to evaluate programs with $\text{\textit{length}} \ll a$ and $\text{\textit{nesting}} \ll b$.
We evaluate the following curriculum learning strategies:

{\bf No curriculum learning (\textit{baseline})}
The baseline approach does not use curriculum learning. 
This means that we generate all the training samples
with $\text{\textit{length}}=a$ and $\text{\textit{nesting}}=b$. 
This strategy is the most ``sound'' from statistical perspective, since it is generally 
recommended to make the training distribution identical to test distribution.

{\bf Naive curriculum strategy (\textit{naive})}
We begin with $\text{\textit{length}}=1$ and $\text{\textit{nesting}}=1$. 
Once learning stops making progress on the validation set, we increase \textit{length} by 1.  We repeat this process until its \textit{length} reaches $a$,
in which case we increase \textit{nesting} by one and reset \textit{length} to $1$.
We can also choose to first increase \textit{nesting} and then \textit{length}. However, it does not make
a noticeable difference in performance. We skip this option in the rest of paper, 
and increase \textit{length} first in all our experiments. This strategy is has been examined in previous work 
on curriculum learning \citep{bengio2009curriculum}. However, we show that sometimes it gives even worse performance than \textit{baseline}.

{\bf Mixed strategy (\textit{mix})}
To generate a random sample, we first pick a random \textit{length} from $[1, a]$ and a random \textit{nesting} from $[1, b]$ 
independently for every sample.
The Mixed strategy uses a balanced mixture of easy and difficult examples, so at every point during training,
a sizable fraction of the training samples will have the appropriate difficulty for the LSTM. 

{\bf Combining the mixed strategy with naive curriculum strategy (\textit{combined})}
This strategy combines the \textit{mix} strategy with the \textit{naive} strategy.  In this approach,
every training case is obtained either by the \textit{naive} strategy or by the \textit{mix} strategy.
As a result, the \textit{combined} strategy always exposes the network at least to some difficult examples,
which is the key way in which it differs from the \textit{naive} curriculum strategy.  
We noticed that it always outperformed the \textit{naive} strategy and would generally (but not always) outperform
the \textit{mix} strategy. We explain why
our new curriculum learning strategies outperform the \textit{naive} curriculum strategy
in Section \ref{sec:hypothesis}.

We evaluate these four strategies on the program evaluation task
(Section \ref{sec:programs_res}) and on the memorization task (Section \ref{sec:copy_res}).

\section{LSTM}
\vspace{-2mm}
\label{sec:lstm}

In this section we briefly describe the deep LSTM (Section \ref{sec:lstm}).
All vectors are $n$-dimensional unless explicitly stated otherwise.  Let $h^l_t
\in \mathbb{R}^{n}$ be a hidden state in layer $l$ in timestep
$t$. Let $T_{n,m}:\mathbb{R}^{n} \rightarrow \mathbb{R}^{m}$
be a biased linear mapping ($x\to Wx + b$ for some $W$ and $b$).
We let $\odot$ be element-wise multiplication and let $h^0_t$ be the
input to the deep LSTM at timestep $t$.  We use the activations at the top layer
$L$ (namely $h^{L}_t$) to predict $y_t$
where $L$ is the depth of our LSTM. 

The structure of the LSTM allows it to train on problems with long term
dependencies relatively easily.  The
``long term'' memory is stored in a vector of \emph{memory cells}
$c^l_t \in \mathbb{R}^n$.  Although many LSTM architectures
differ slightly in their connectivity structure and activation functions,
all LSTM architectures have additive memory cells that make it easy to learn
to store information for long periods of time. 
We used an LSTM described by the following equations (from \citet{graves2013speech}):
\begin{align*}
&\text{LSTM} : h^{l-1}_t, h^l_{t-1}, c^l_{t - 1} \rightarrow h^l_t, c^l_t\\
&\begin{pmatrix}i\\f\\o\\g\end{pmatrix} =
  \begin{pmatrix}\mathrm{sigm}\\\mathrm{sigm}\\\mathrm{sigm}\\\tanh\end{pmatrix}
  T_{2n,4n}\begin{pmatrix}h^{l - 1}_t\\h^l_{t-1}\end{pmatrix}\\
&c^l_t = f \odot c^l_{t-1} + i \odot g\\
&h^l_t = o \odot \tanh(c^l_t)\\
\end{align*}
\vspace{-10mm}

\section{Experiments}
\vspace{-2mm}

In this section, we report the results of our curriculum learning strategies on 
the program evaluation and memorization tasks. In both experiments, we used the same LSTM architecture.

Our LSTM has two layers and is unrolled for $50$ steps in both experiments. 
It has $400$ cells per layer and its parameters are
initialized uniformly in $[-0.08, 0.08]$. This gives total $\sim 2.5$M parameters.
We initialize the hidden states to zero.  We then use the final hidden states of
the current minibatch as the initial hidden state of the subsequent minibatch.  Thus it is possible
that a program and its output could be separated across different minibatches. 
The size of minibatch is $100$.  We constrain the norm of
the gradients (normalized by minibatch size) to be no greater than $5$ \citep{mikolov2010recurrent}. 
We keep the learning rate equal to $0.5$ until we reach the target 
\textit{length} and \textit{nesting} (we only vary the \textit{length}, i.e., 
the number of digits, in the memorization task).

After reaching the target accuracy ($95\%$) we decrease the learning rate by $0.8$.
We keep the learning rate on the same level until there is no improvement on the training set.
We decrease it again, when there is no improvement on training set. 
The only difference between experiments is the termination criteria. For the program
output prediction, we stop when learning rate becomes smaller than $0.001$.
For copying task, we stop training after $20$ epochs, where each epoch has $0.5$M samples.

We begin training with $\text{\textit{length}}=1$ and $\text{\textit{nesting}}=1$ (or 
\textit{length}=1 for the memorization task).  We ensure that the training, validation,
and test sets are disjoint. It is achieved computing the hash value of each sample and taking it modulo 3.

{\bf Important note on error rates:}
We use teacher forcing  when we compute the accuracy of our LSTMs.   
That is, when predicting the $i$-th digit of the target, the LSTM 
is provided with the \emph{correct} first $i-1$ digits
of the target.  This is different from using the LSTM to generate the entire
output on its own, as done by \cite{sutskever2014sequence}, which would almost surely result in lower numerical accuracies.  To
help make intuitive sense of our results, we present a large number of test cases and the
outputs computed by the LSTM, albeit with teacher forcing.  

\subsection{Results on Program Evaluation}
\label{sec:programs_res}
\vspace{-2mm}

\begin{figure}[h]
  \centering
  \includegraphics[width=0.26\linewidth]{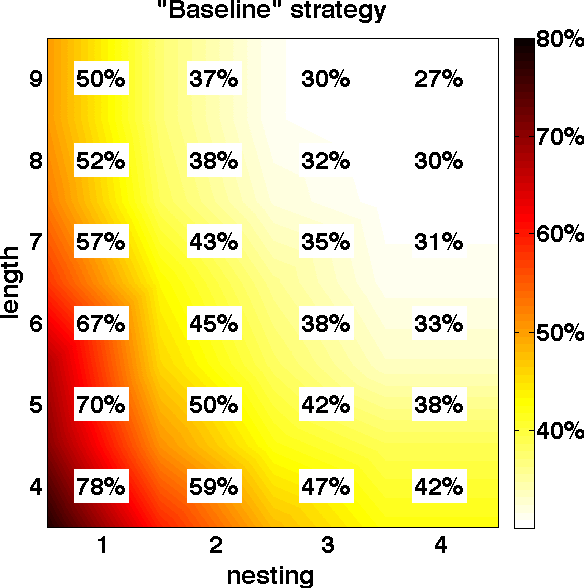}
  \includegraphics[width=0.26\linewidth]{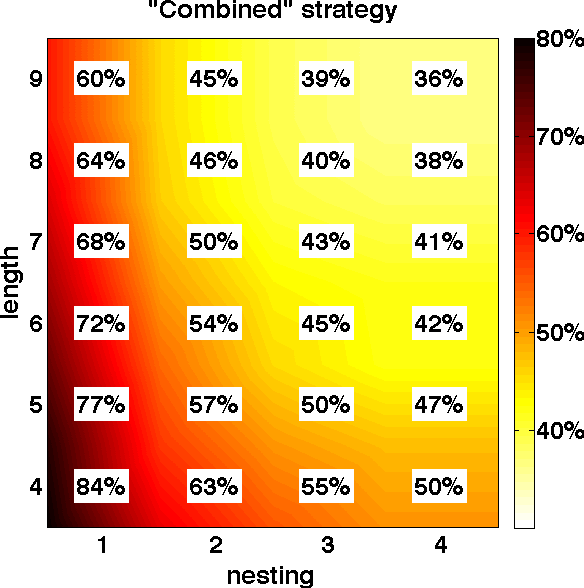}
  \caption{Absolute prediction accuracy of the \textit{baseline} strategy
    and of the \textit{combined} strategy (see Section \ref{sec:curriculum}) on the program evaluation task.
    Deeper nesting and longer integers make the task more difficult. Overall,
    the \textit{combined} strategy outperformed the \textit{baseline}
    strategy in every setting.
  }
  \label{fig:programs_abs}
\end{figure}

\begin{figure}[h]
  \centering
  \subfigure{
    \includegraphics[width=0.26\linewidth]{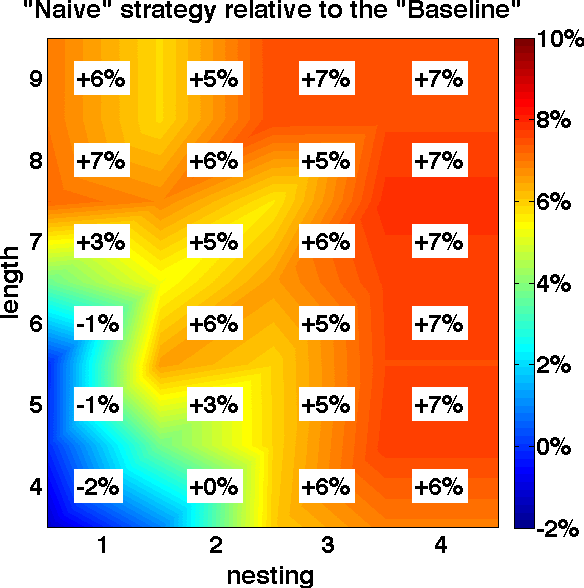}
    \includegraphics[width=0.26\linewidth]{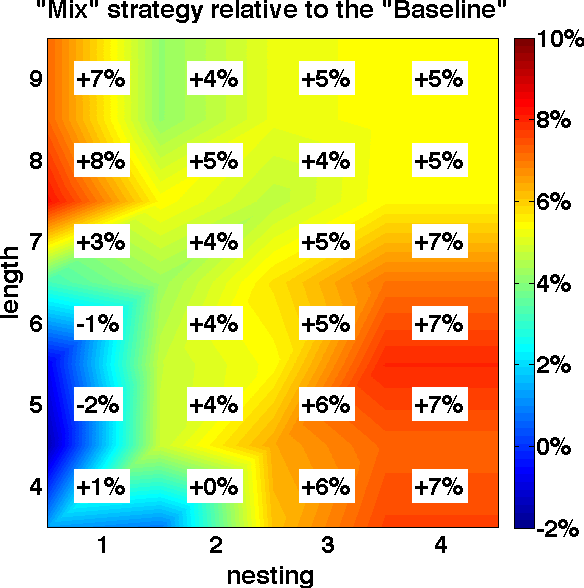}
    \includegraphics[width=0.26\linewidth]{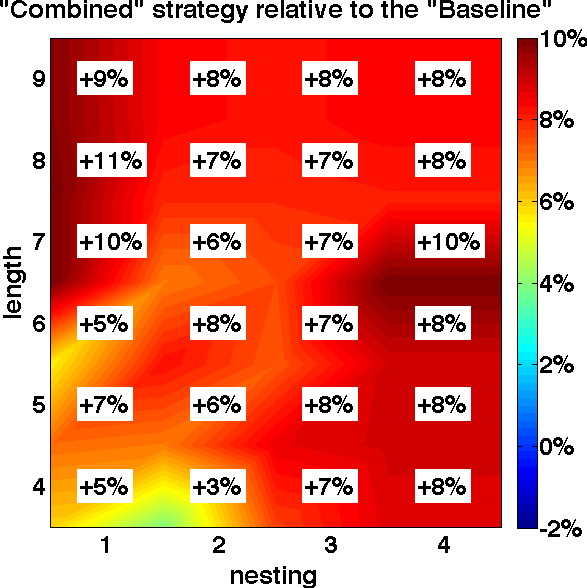}
  }
  \caption{Relative prediction accuracy of the different strategies with respect to 
    the \textit{baseline} strategy. The \textit{Naive} curriculum strategy  
was found to sometime perform {\bf worse} than \textit{baseline}. A possible explanation is provided in
Section \ref{sec:hypothesis}. The \textit{combined} strategy outperforms all other strategies in every
configuration on program evaluation.}
  \label{fig:programs_rel}
\end{figure}

We train our LSTMs using the four strategies described in Section \ref{sec:curriculum}:
\begin{itemize}
  \item No curriculum learning (\textit{baseline}),
  \item Naive curriculum strategy (\textit{naive})
  \item Mixed strategy  (\textit{mix}), and
  \item Combined strategy (\textit{combined}).
\end{itemize}
Figure \ref{fig:programs_abs} shows the absolute performance of the \textit{baseline} strategy
(training on the original target distribution), and
of the best performing strategy, \textit{combined}. Moreover, Figure 
\ref{fig:programs_rel} shows the performance of the three curriculum strategies relative to
\textit{baseline}. Finally, we provide several example 
predictions on test data in the supplementary materials.
The accuracy of a random predictor would be $\sim 8.3\%$, since there are 
$12$ possible output symbols.  \vspace{-4mm}
\newline

\vspace{-3mm}

\subsection{Results on the Addition Task}
\vspace{-2mm}
\label{sec:add_res}

\begin{figure}
\centerline{
  \includegraphics[width=0.46\linewidth]{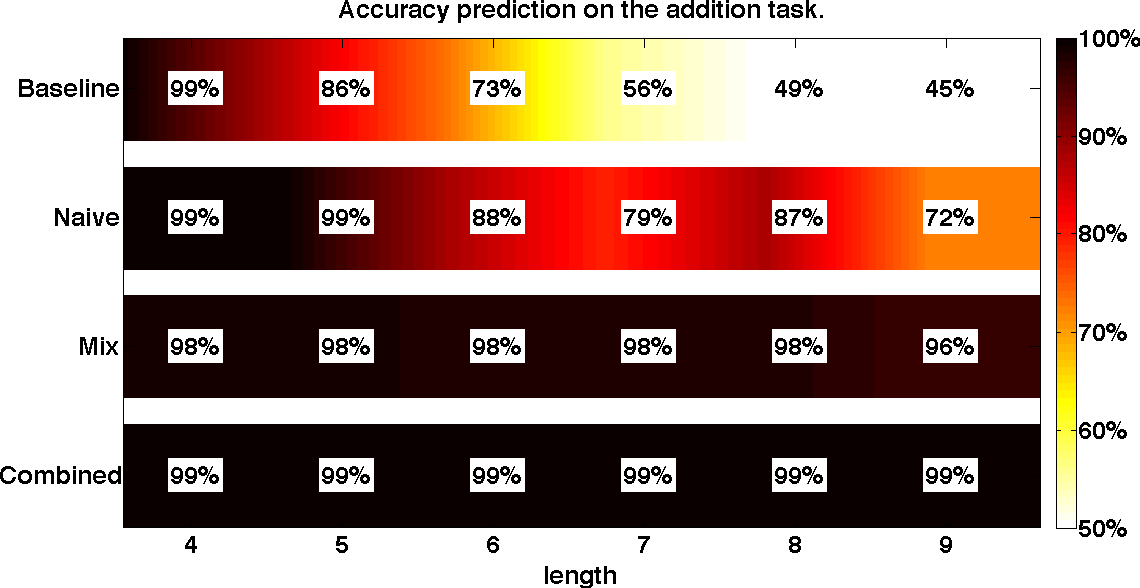}}
\vspace{-5mm}
  \caption{\footnotesize{The effect of curriculum strategies on the addition task.}}
  \label{fig:cur_addition}
\end{figure}

Figure \ref{fig:cur_addition} presents the accuracy achieved by the LSTM with the various  
curriculum strategies on the addition task.  Remarkably, the \textit{combined} curriculum strategy resulted in
99\% accuracy on the addition of 9-digit long numbers, which is a massive improvement over 
the \textit{naive} curriculum. 

\subsection{Results on the Memorization Task}
\vspace{-2mm}

\begin{figure}[t!]
\centerline{
  \subfigure{
    \includegraphics[width=0.46\linewidth]{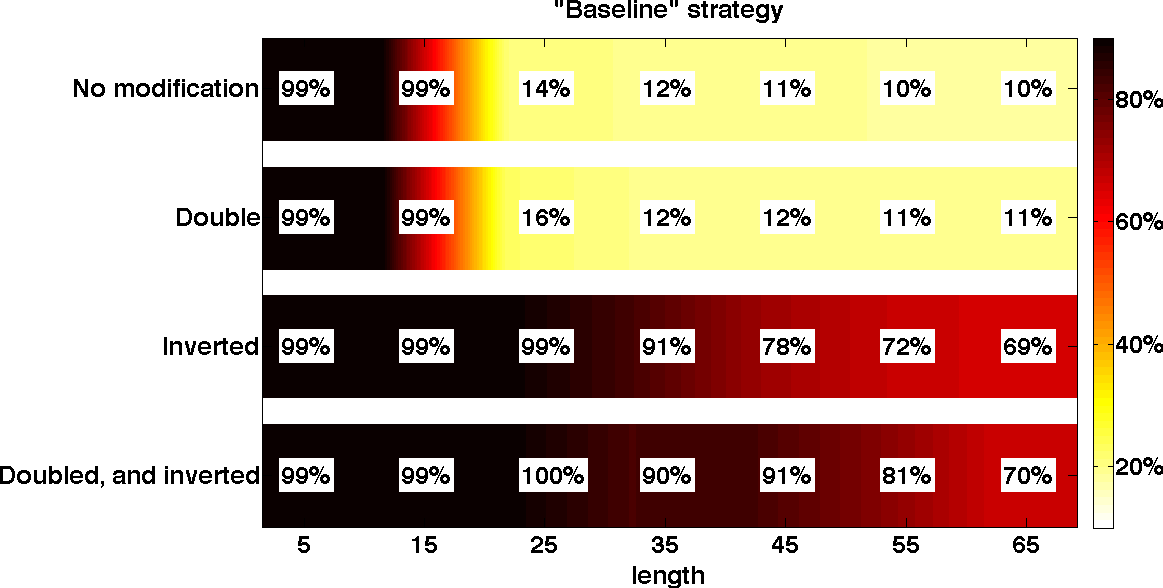}
    \hspace{3mm}
    \includegraphics[width=0.46\linewidth]{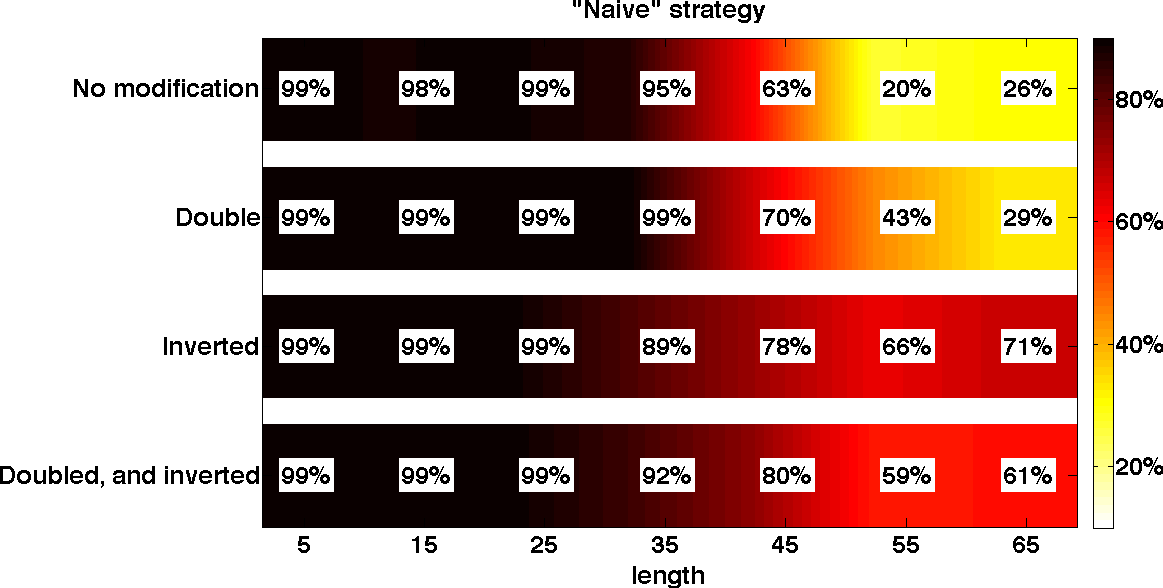}
  }
}
\centerline{
  \subfigure{
    \includegraphics[width=0.46\linewidth]{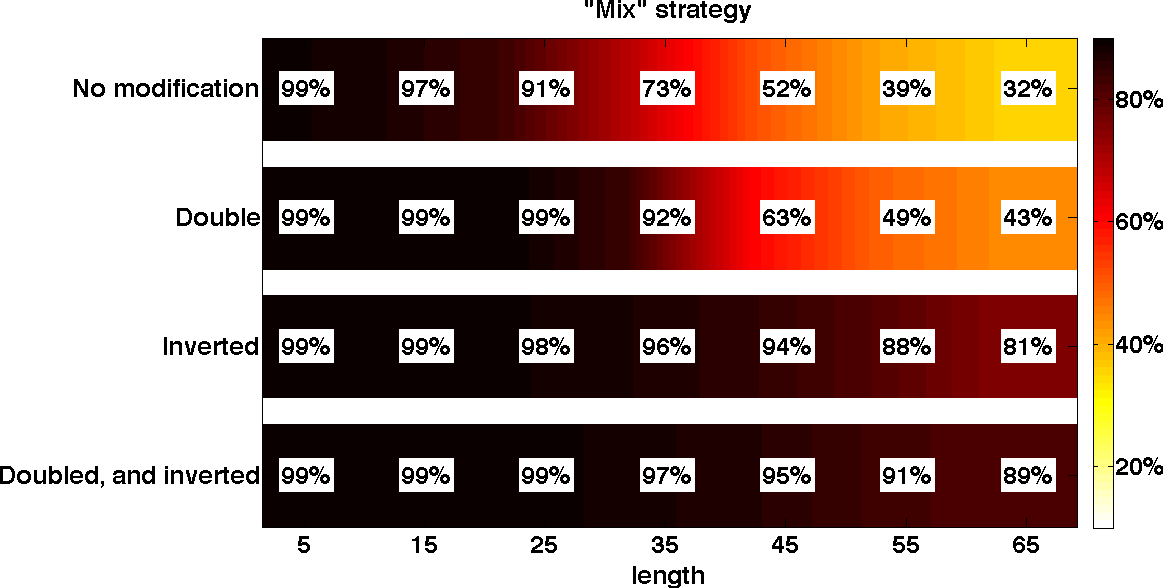}
    \hspace{3mm}
    \includegraphics[width=0.46\linewidth]{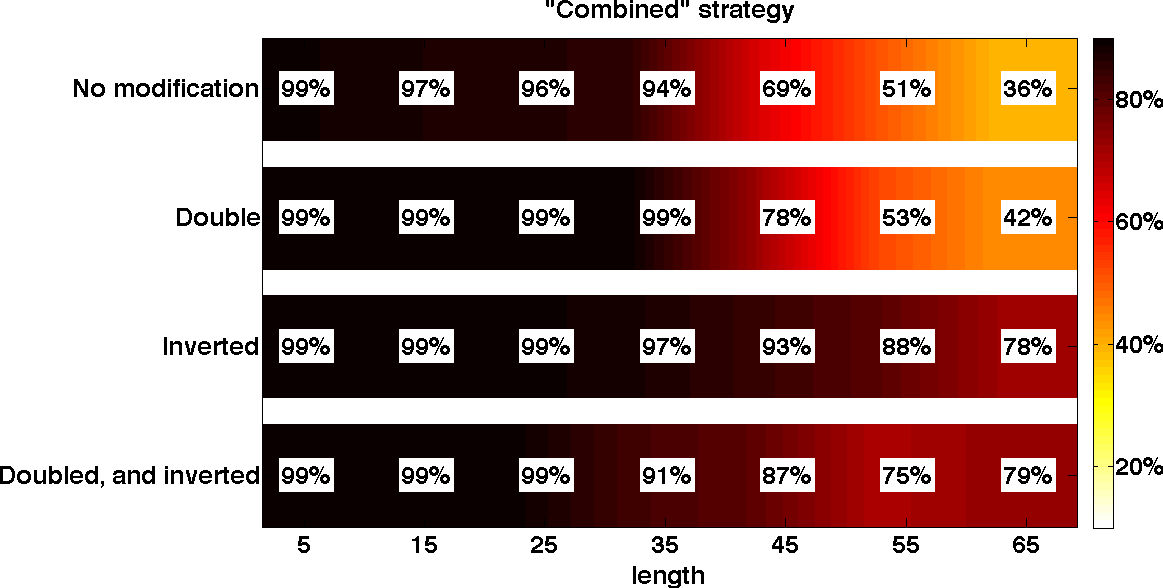}
  }
}
  \caption{Prediction accuracy on the memorization task for the four curriculum strategies.
           The input length ranges from $5$ to $65$ digits. Every strategy is evaluated with the following
           $4$ input modification schemes: no modification; input inversion; input doubling;  and
           input doubling and inversion. The training time was not limited;  the network was trained till convergence.
}
  \label{fig:copy_abs}
\end{figure}

\begin{figure}
\centerline{
\subfigure{
\includegraphics[width=0.47\linewidth]{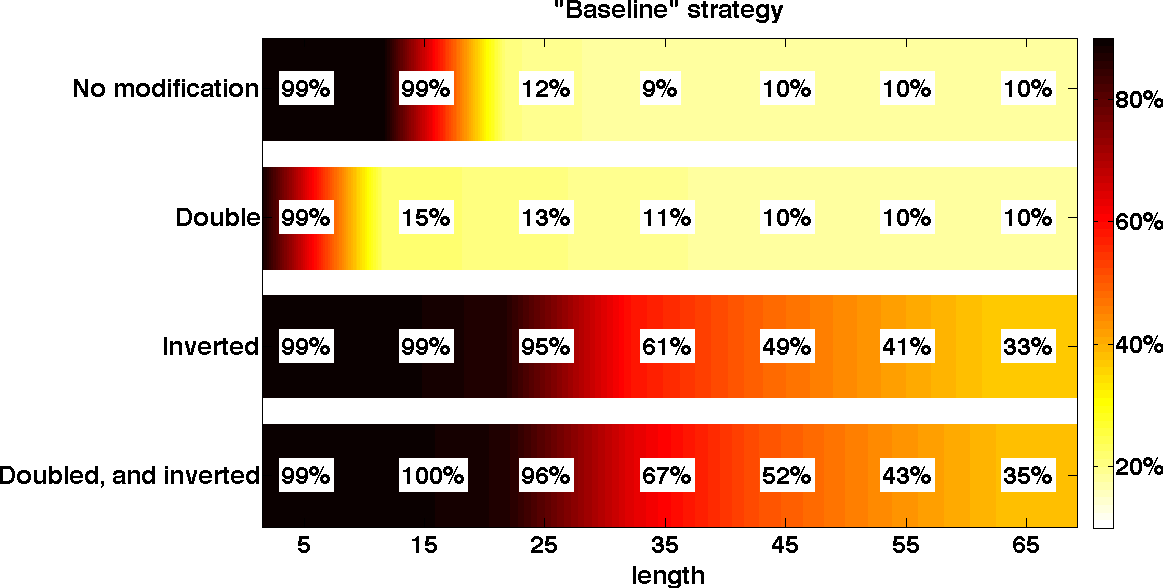}
  }
  \subfigure{
\includegraphics[width=0.47\linewidth]{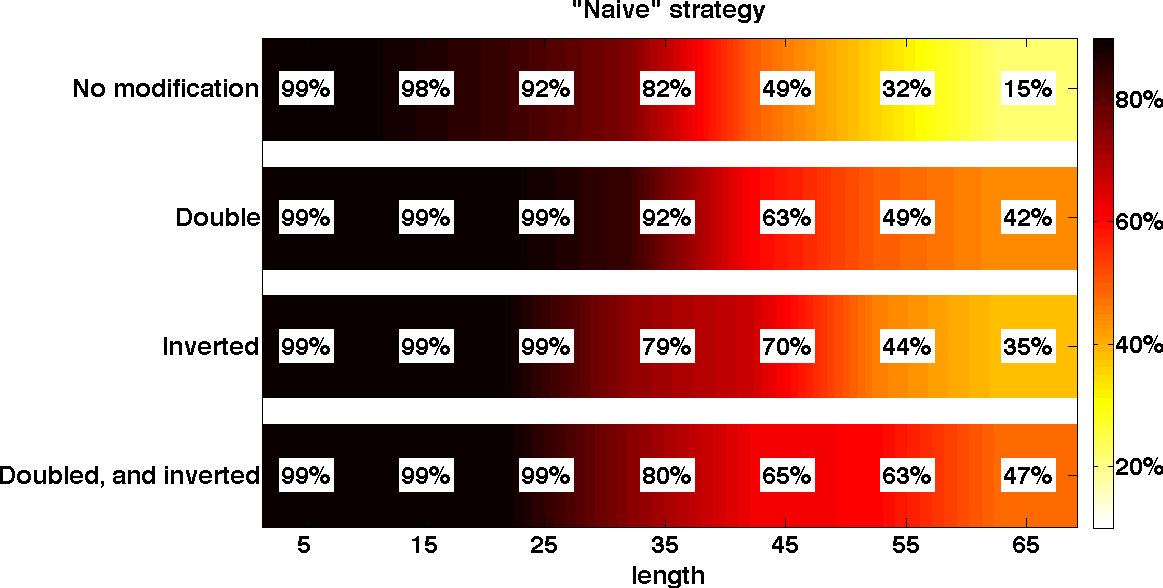}
}
}
\centerline{
\subfigure{
\includegraphics[width=0.47\linewidth]{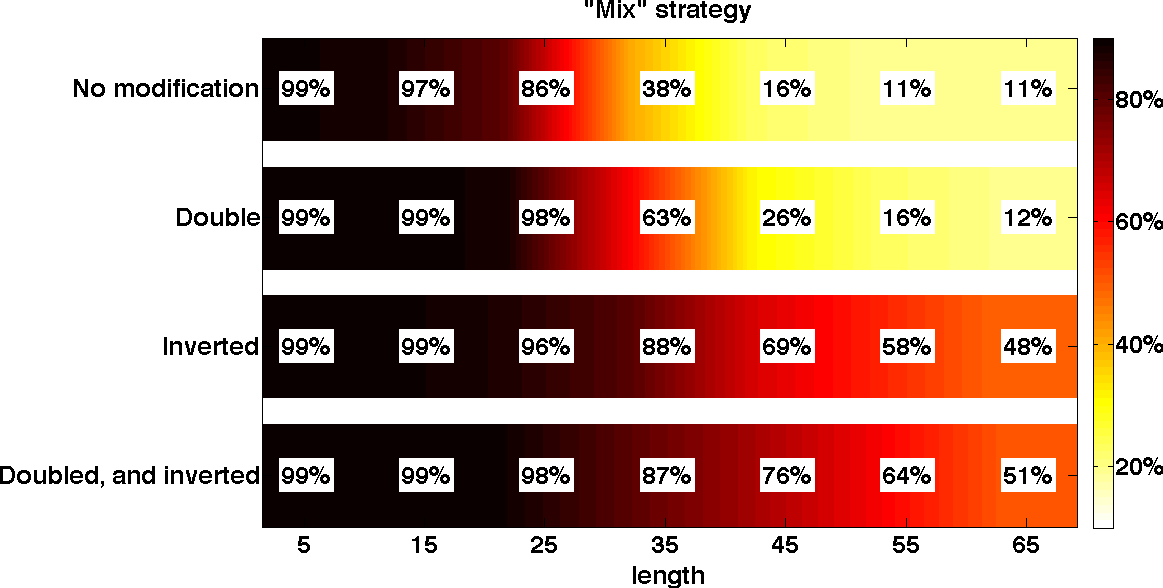}
  }
  \subfigure{
\includegraphics[width=0.47\linewidth]{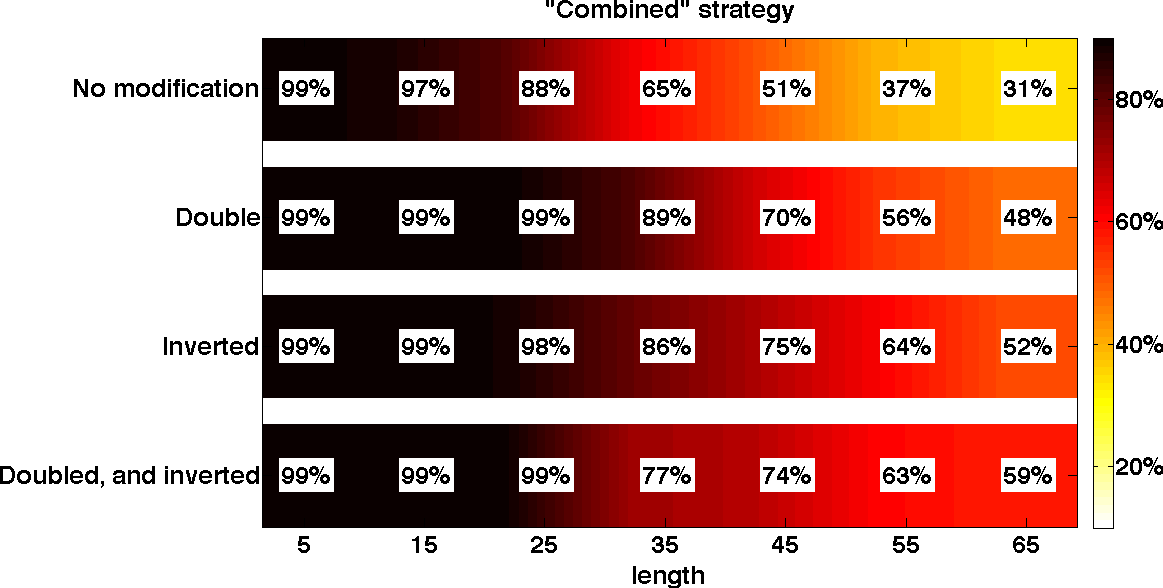}
}
}

\caption{
  Prediction accuracy on the memorization task for the four curriculum strategies.
  The input length ranges from $5$ to $65$ digits. Every strategy is evaluated with the following
  $4$ input modification schemes: no modification; input inversion; input doubling;  and
  input doubling and inversion. The training time is limited to $20$ epochs.  
}
  \label{fig:copy_abs_20}
\end{figure}

Recall that the goal of the memorization task is to read a sequence of digits into the hidden state and then
to reconstruct it from the hidden state. Namely, given an input such as $123456789$, the goal 
is to produce the output $123456789$. 
The model processes the input one input character at the time and has to reconstruct the output only after 
loading the entire input into its memory. This task provides insight into the LSTM's 
ability to learn to remember.  We have
evaluated our model on sequences of lengths ranging from $5$ to $65$. 
We use the four curriculum strategies of Section \ref{sec:curriculum}. In addition,
we investigate two strategies to modify the input which increase performance:
\begin{itemize}
  \item Inverting input \citep{sutskever2014sequence}
  \item Doubling Input
\end{itemize}
Both strategies are described in Section \ref{sec:copy}. Figure \ref{fig:copy_abs} shows the absolute performance
of the \textit{baseline} strategy and of the \textit{combined} strategy. 
This Figure shows the performance at convergence.  We further present in Supplementary material (Section \ref{sec:sup}) results after $20$ epochs (Figure \ref{fig:copy_abs_20}).

For this task, the \textit{combined} strategy no longer outperforms the \textit{mixed} strategy  
in every experimental setting, although both strategies are always better than using no curriculum and the \textit{naive} curriculum strategy.
Each graph contains $4$ settings, which correspond to the possible combinations
of input inversion and input doubling.  The result clearly shows that the simultaneously doubling and reversing the input
achieves the best results.  Random guessing 
would achieve an accuracy of $\sim 9\%$, since there are $11$ possible output symbols.
\label{sec:copy_res}

\section{Hidden State Allocation Hypothesis}
\vspace{-1mm}
\label{sec:hypothesis}
Our experimental results suggest that a proper curriculum learning strategy 
is critical for achieving good performance on very hard problems where conventional stochastic 
gradient descent (SGD) performs poorly. The results on both of our problems
 (Sections \ref{sec:copy_res} and \ref{sec:programs_res}) show
that the \textit{combined} strategy is better than all other curriculum strategies, including both \textit{naive} curriculum learning,
and training on the target distribution. We have a plausible explanation for why this is the case.

It seems natural to train models with examples of increasing difficulty. This way the models have a chance
to learn the correct intermediate concepts, 
and then utilize them for the more difficult problem instances. 
Otherwise, learning the full task might be just too difficult for SGD from a random initialization.
This explanation has been proposed in previous work on curriculum learning \cite{bengio2009curriculum}.  
However, based the on empirical results, the \textit{naive} strategy of curriculum learning 
can sometimes be worse than learning with the target distribution.

In our tasks, the neural network has to perform a lot of memorization. The easier examples usually require less memorization
than the hard examples. For instance, in order to add two $5$-digit numbers, one has to remember at least $5$
digits before producing any output. The best way to accurately memorize $5$ numbers could be to spread them
over the entire hidden state / memory cell (i.e., use a \emph{distributed representation}). Indeed, the 
network has no incentive to utilize only a fraction of its state, and it is always better to make use of its entire memory capacity.
This implies that the harder examples would require a restructuring of its memory patterns. 
It would need to contract its representations of $5$ digit numbers in order to free space for the $6$-th number.
This process of memory pattern restructuring might be difficult to implement, so it could be the reason for the
sometimes poor performance of the \textit{naive} curriculum learning strategy relative to \textit{baseline}.

The \textit{combined} strategy reduces the need to restructure the memory patterns. The \textit{combined} strategy is a combination
of the \textit{naive} curriculum strategy and of the \textit{mix} strategy, which is a mixture of examples of all difficulties. The examples 
produced by the \textit{naive} curriculum strategy
help to learn the intermediate input-output mapping, which is useful for solving the target task, while the extra samples 
from the \textit{mix} strategy prevent the network from utilizing all the memory on the easy examples, thus  eliminating
the need to restructure its memory patterns. 

\section{Critique}
\vspace{-1mm}

Perfect prediction of program output requires a complete understanding of all operands and concepts, and of the precise
way in which they are combined.
However, imperfect prediction might be achieved in a multitude of ways, and could heavily rely
on memorization, without a genuine understanding of the underlying concepts. For instance, perfect addition  
is relatively intricate, as the LSTM needs to know the order of numbers and to correctly compute the carry. 

There are many alternatives to the addition algorithm if perfect output is not required. 
For instance, one can perform element-wise addition, and as long as there is no carry then
the output would be perfectly correct. Another alternative, which requires more memory, but is also more
simpler, is to memorize all results of addition for $2$ digit numbers. Then multi-digit addition can
be broken down to multiple $2$-digits additions element-wise. Once again, such an algorithm
would have a reasonably high prediction accuracy, although it would be far from correct.

We do not know how heavily our model relies on memorization and how far the learned algorithm
is from the actual, correct algorithm. This could be tested by creating a big discrepancy
between the training and test data, but in this work, the training and the test distributions 
are the same. We plan to examine how well our models
would generalize on very different new examples in future work.

\section{Discussion}
\vspace{-1mm}

We have shown that it is possible to learn to evaluate programs with limited prior knowledge. 
This work demonstrate the power and expressiveness of sequence-to-sequence LSTMs.  We also showed that correct curriculum
learning is crucial for achieving good results on very difficult tasks that cannot be optimized
with standard SGD.  We also found that the general method of doubling the input reliably improves the performance of sequence-to-sequence LSTMs.

Our results are encouraging but they leave many questions open.  For example, we are not able to evaluate arbitrary programs
(e.g., ones that run in more than $\BigO{n}$ time). This cannot be achieved with conventional RNNs or LSTMs due to 
their runtime restrictions. We also do not know the optimal curriculum learning strategy. To
understand it, it may be necessary to identify the training samples that are most beneficial to the model.

\footnotesize{
\section{Acknowledgments}

We wish to thank Oriol Vinyals for useful discussions, and to 
Koray Kavukcuoglu for help during code development. 
Moreover, we wish to acknowledge Marc'Aurelio Ranzato for useful comments on the first version of the paper.   
Some chunks of our code origin from Google Deepmind repository. We thank to unknown developers 
of LSTM function, and auxiliary functions.

}
\bibliography{bibliography}
\bibliographystyle{icml2014}

\clearpage
\section*{Supplementary material}
\label{sec:sup}

\begin{algorithm}[h]
\footnotesize{
\label{alg:prog_gen}
\texttt{
\begin{algorithmic}
\STATE {\bfseries Input:} length, nesting
\STATE stack = EmptyStack()
\STATE Operations = {Addition, Subtraction, Multiplication, If-Statement, For-Loop, Variable Assignment} 
\FOR{$i=1$ {\bfseries to} nesting}
  \STATE Operation = a random operation from Operations
  \STATE Values = List
  \STATE Code = List
  \FOR{params in Operation.params}
    \IF {not empty stack and Uniform(1) $> 0.5$}
      \STATE value, code = stack.pop()
    \ELSE
    \STATE value = random.int($10^{length}$)
      \STATE code = toString(value)
    \ENDIF
    \STATE values.append(value)
    \STATE code.append(code)
  \ENDFOR
  \STATE new\_value= Operation.evaluate(values)
  \STATE new\_code = Operation.generate\_code(codes)
  \STATE stack.push((new\_value, new\_code))
\ENDFOR
\STATE final\_value, final\_code = stack.pop()
\STATE datasets = {training, validation, testing}
\STATE idx = hash(final\_code) modulo 3
\STATE datasets[idx].add((final\_value, final\_code))
\end{algorithmic}
}
}
\caption{
Pseudocode of the algorithm used to generate the distribution over the python program.  
Programs produced by this algorithm are guaranteed to never have dead code.
The type of the sample (train, test, or validation) is determined by its hash modulo 3. 
}
\end{algorithm}

\section{Additional Results on the Memorization Problem}

We present the algorithm for generating the training cases, and present an extensive
qualitative evaluation of the samples and the kinds of predictions 
made by the trained LSTMs.  

We emphasize that these predictions rely on \emph{teacher forcing}.  That is,
even if the LSTM made an incorrect prediction in the $i$-th  output digit, 
the LSTM will be provided as input the \emph{correct} $i$-th output digit 
for predicting the $i+1$-th digit.  While teacher forcing has no effect  
whenever the LSTM makes no errors at all, a sample that makes an early
error and gets   the remainder of the digits correctly needs to be interpreted
with care.

\section{Qualitative evaluation of the curriculum strategies}

\vspace{0.3cm}\subsection{Examples of program evaluation prediction. Length = 4, Nesting = 1}\vspace{0.3cm}

\begin{mdframed}\vspace{0.07cm}
{\bf Input:}
\begin{lstlisting}
print(6652).
\end{lstlisting}
\begin{tabular}{ l l }
{\bf Target:} & 6652. \\
{\bf "Baseline" prediction:} & {\color[rgb]{0.31, 0.49, 0.16} 6}{\color[rgb]{0.31, 0.49, 0.16} 6}{\color[rgb]{0.31, 0.49, 0.16} 5}{\color[rgb]{0.31, 0.49, 0.16} 2}{\color[rgb]{0.31, 0.49, 0.16} .}\\
{\bf "Naive" prediction:} & {\color[rgb]{0.31, 0.49, 0.16} 6}{\color[rgb]{0.31, 0.49, 0.16} 6}{\color[rgb]{0.31, 0.49, 0.16} 5}{\color[rgb]{0.31, 0.49, 0.16} 2}{\color[rgb]{0.31, 0.49, 0.16} .}\\
{\bf "Mix" prediction:} & {\color[rgb]{0.31, 0.49, 0.16} 6}{\color[rgb]{0.31, 0.49, 0.16} 6}{\color[rgb]{0.31, 0.49, 0.16} 5}{\color[rgb]{0.31, 0.49, 0.16} 2}{\color[rgb]{0.31, 0.49, 0.16} .}\\
{\bf "Combined" prediction:} & {\color[rgb]{0.31, 0.49, 0.16} 6}{\color[rgb]{0.31, 0.49, 0.16} 6}{\color[rgb]{0.31, 0.49, 0.16} 5}{\color[rgb]{0.31, 0.49, 0.16} 2}{\color[rgb]{0.31, 0.49, 0.16} .}\\
\end{tabular}

\end{mdframed}\vspace{0.07cm}
\vspace{-0.1cm}\begin{mdframed}\vspace{0.07cm}
{\bf Input:}
\begin{lstlisting}
print((5997-738)).
\end{lstlisting}
\begin{tabular}{ l l }
{\bf Target:} & 5259. \\
{\bf "Baseline" prediction:} & {\color[rgb]{0.31, 0.49, 0.16} 5}{\color[rgb]{0.8,0.2,0.2} 1}{\color[rgb]{0.8,0.2,0.2} 0}{\color[rgb]{0.8,0.2,0.2} 1}{\color[rgb]{0.31, 0.49, 0.16} .}\\
{\bf "Naive" prediction:} & {\color[rgb]{0.31, 0.49, 0.16} 5}{\color[rgb]{0.8,0.2,0.2} 1}{\color[rgb]{0.8,0.2,0.2} 0}{\color[rgb]{0.8,0.2,0.2} 1}{\color[rgb]{0.31, 0.49, 0.16} .}\\
{\bf "Mix" prediction:} & {\color[rgb]{0.31, 0.49, 0.16} 5}{\color[rgb]{0.31, 0.49, 0.16} 2}{\color[rgb]{0.8,0.2,0.2} 4}{\color[rgb]{0.31, 0.49, 0.16} 9}{\color[rgb]{0.31, 0.49, 0.16} .}\\
{\bf "Combined" prediction:} & {\color[rgb]{0.31, 0.49, 0.16} 5}{\color[rgb]{0.31, 0.49, 0.16} 2}{\color[rgb]{0.8,0.2,0.2} 2}{\color[rgb]{0.31, 0.49, 0.16} 9}{\color[rgb]{0.31, 0.49, 0.16} .}\\
\end{tabular}

\end{mdframed}\vspace{0.07cm}
\vspace{-0.1cm}\begin{mdframed}\vspace{0.07cm}
{\bf Input:}
\begin{lstlisting}
print((16*3071)).
\end{lstlisting}
\begin{tabular}{ l l }
{\bf Target:} & 49136. \\
{\bf "Baseline" prediction:} & {\color[rgb]{0.31, 0.49, 0.16} 4}{\color[rgb]{0.31, 0.49, 0.16} 9}{\color[rgb]{0.8,0.2,0.2} 3}{\color[rgb]{0.31, 0.49, 0.16} 3}{\color[rgb]{0.31, 0.49, 0.16} 6}{\color[rgb]{0.31, 0.49, 0.16} .}\\
{\bf "Naive" prediction:} & {\color[rgb]{0.31, 0.49, 0.16} 4}{\color[rgb]{0.8,0.2,0.2} 8}{\color[rgb]{0.8,0.2,0.2} 6}{\color[rgb]{0.8,0.2,0.2} 7}{\color[rgb]{0.31, 0.49, 0.16} 6}{\color[rgb]{0.31, 0.49, 0.16} .}\\
{\bf "Mix" prediction:} & {\color[rgb]{0.8,0.2,0.2} 5}{\color[rgb]{0.8,0.2,0.2} 7}{\color[rgb]{0.8,0.2,0.2} 0}{\color[rgb]{0.8,0.2,0.2} 2}{\color[rgb]{0.31, 0.49, 0.16} 6}{\color[rgb]{0.31, 0.49, 0.16} .}\\
{\bf "Combined" prediction:} & {\color[rgb]{0.31, 0.49, 0.16} 4}{\color[rgb]{0.31, 0.49, 0.16} 9}{\color[rgb]{0.8,0.2,0.2} 6}{\color[rgb]{0.8,0.2,0.2} 2}{\color[rgb]{0.31, 0.49, 0.16} 6}{\color[rgb]{0.31, 0.49, 0.16} .}\\
\end{tabular}

\end{mdframed}\vspace{0.07cm}
\vspace{-0.1cm}\begin{mdframed}\vspace{0.07cm}
{\bf Input:}
\begin{lstlisting}
c=2060;
print((c-4387)).
\end{lstlisting}
\begin{tabular}{ l l }
{\bf Target:} & -2327. \\
{\bf "Baseline" prediction:} & {\color[rgb]{0.31, 0.49, 0.16} -}{\color[rgb]{0.31, 0.49, 0.16} 2}{\color[rgb]{0.31, 0.49, 0.16} 3}{\color[rgb]{0.31, 0.49, 0.16} 2}{\color[rgb]{0.8,0.2,0.2} 0}{\color[rgb]{0.31, 0.49, 0.16} .}\\
{\bf "Naive" prediction:} & {\color[rgb]{0.31, 0.49, 0.16} -}{\color[rgb]{0.31, 0.49, 0.16} 2}{\color[rgb]{0.8,0.2,0.2} 2}{\color[rgb]{0.8,0.2,0.2} 0}{\color[rgb]{0.8,0.2,0.2} 1}{\color[rgb]{0.31, 0.49, 0.16} .}\\
{\bf "Mix" prediction:} & {\color[rgb]{0.31, 0.49, 0.16} -}{\color[rgb]{0.31, 0.49, 0.16} 2}{\color[rgb]{0.31, 0.49, 0.16} 3}{\color[rgb]{0.8,0.2,0.2} 7}{\color[rgb]{0.31, 0.49, 0.16} 7}{\color[rgb]{0.31, 0.49, 0.16} .}\\
{\bf "Combined" prediction:} & {\color[rgb]{0.31, 0.49, 0.16} -}{\color[rgb]{0.31, 0.49, 0.16} 2}{\color[rgb]{0.31, 0.49, 0.16} 3}{\color[rgb]{0.8,0.2,0.2} 1}{\color[rgb]{0.31, 0.49, 0.16} 7}{\color[rgb]{0.31, 0.49, 0.16} .}\\
\end{tabular}

\end{mdframed}\vspace{0.07cm}
\vspace{-0.1cm}\begin{mdframed}\vspace{0.07cm}
{\bf Input:}
\begin{lstlisting}
print((2*5172)).
\end{lstlisting}
\begin{tabular}{ l l }
{\bf Target:} & 10344. \\
{\bf "Baseline" prediction:} & {\color[rgb]{0.31, 0.49, 0.16} 1}{\color[rgb]{0.31, 0.49, 0.16} 0}{\color[rgb]{0.31, 0.49, 0.16} 3}{\color[rgb]{0.31, 0.49, 0.16} 4}{\color[rgb]{0.31, 0.49, 0.16} 4}{\color[rgb]{0.31, 0.49, 0.16} .}\\
{\bf "Naive" prediction:} & {\color[rgb]{0.31, 0.49, 0.16} 1}{\color[rgb]{0.31, 0.49, 0.16} 0}{\color[rgb]{0.31, 0.49, 0.16} 3}{\color[rgb]{0.8,0.2,0.2} 2}{\color[rgb]{0.31, 0.49, 0.16} 4}{\color[rgb]{0.31, 0.49, 0.16} .}\\
{\bf "Mix" prediction:} & {\color[rgb]{0.31, 0.49, 0.16} 1}{\color[rgb]{0.31, 0.49, 0.16} 0}{\color[rgb]{0.31, 0.49, 0.16} 3}{\color[rgb]{0.31, 0.49, 0.16} 4}{\color[rgb]{0.31, 0.49, 0.16} 4}{\color[rgb]{0.31, 0.49, 0.16} .}\\
{\bf "Combined" prediction:} & {\color[rgb]{0.31, 0.49, 0.16} 1}{\color[rgb]{0.31, 0.49, 0.16} 0}{\color[rgb]{0.31, 0.49, 0.16} 3}{\color[rgb]{0.31, 0.49, 0.16} 4}{\color[rgb]{0.31, 0.49, 0.16} 4}{\color[rgb]{0.31, 0.49, 0.16} .}\\
\end{tabular}

\end{mdframed}\vspace{0.07cm}
\vspace{-0.1cm}\begin{mdframed}\vspace{0.07cm}
{\bf Input:}
\begin{lstlisting}
print((9891-4715)).
\end{lstlisting}
\begin{tabular}{ l l }
{\bf Target:} & 5176. \\
{\bf "Baseline" prediction:} & {\color[rgb]{0.31, 0.49, 0.16} 5}{\color[rgb]{0.31, 0.49, 0.16} 1}{\color[rgb]{0.8,0.2,0.2} 9}{\color[rgb]{0.31, 0.49, 0.16} 6}{\color[rgb]{0.31, 0.49, 0.16} .}\\
{\bf "Naive" prediction:} & {\color[rgb]{0.31, 0.49, 0.16} 5}{\color[rgb]{0.31, 0.49, 0.16} 1}{\color[rgb]{0.8,0.2,0.2} 0}{\color[rgb]{0.8,0.2,0.2} 4}{\color[rgb]{0.31, 0.49, 0.16} .}\\
{\bf "Mix" prediction:} & {\color[rgb]{0.8,0.2,0.2} 4}{\color[rgb]{0.8,0.2,0.2} 2}{\color[rgb]{0.8,0.2,0.2} 4}{\color[rgb]{0.31, 0.49, 0.16} 6}{\color[rgb]{0.31, 0.49, 0.16} .}\\
{\bf "Combined" prediction:} & {\color[rgb]{0.31, 0.49, 0.16} 5}{\color[rgb]{0.31, 0.49, 0.16} 1}{\color[rgb]{0.8,0.2,0.2} 9}{\color[rgb]{0.31, 0.49, 0.16} 6}{\color[rgb]{0.31, 0.49, 0.16} .}\\
\end{tabular}

\end{mdframed}\vspace{0.07cm}
\vspace{-0.1cm}\begin{mdframed}\vspace{0.07cm}
{\bf Input:}
\begin{lstlisting}
print(4849).
\end{lstlisting}
\begin{tabular}{ l l }
{\bf Target:} & 4849. \\
{\bf "Baseline" prediction:} & {\color[rgb]{0.31, 0.49, 0.16} 4}{\color[rgb]{0.31, 0.49, 0.16} 8}{\color[rgb]{0.31, 0.49, 0.16} 4}{\color[rgb]{0.31, 0.49, 0.16} 9}{\color[rgb]{0.31, 0.49, 0.16} .}\\
{\bf "Naive" prediction:} & {\color[rgb]{0.31, 0.49, 0.16} 4}{\color[rgb]{0.31, 0.49, 0.16} 8}{\color[rgb]{0.31, 0.49, 0.16} 4}{\color[rgb]{0.31, 0.49, 0.16} 9}{\color[rgb]{0.31, 0.49, 0.16} .}\\
{\bf "Mix" prediction:} & {\color[rgb]{0.31, 0.49, 0.16} 4}{\color[rgb]{0.31, 0.49, 0.16} 8}{\color[rgb]{0.31, 0.49, 0.16} 4}{\color[rgb]{0.31, 0.49, 0.16} 9}{\color[rgb]{0.31, 0.49, 0.16} .}\\
{\bf "Combined" prediction:} & {\color[rgb]{0.31, 0.49, 0.16} 4}{\color[rgb]{0.31, 0.49, 0.16} 8}{\color[rgb]{0.31, 0.49, 0.16} 4}{\color[rgb]{0.31, 0.49, 0.16} 9}{\color[rgb]{0.31, 0.49, 0.16} .}\\
\end{tabular}

\end{mdframed}\vspace{0.07cm}
\vspace{-0.1cm}\begin{mdframed}\vspace{0.07cm}
{\bf Input:}
\begin{lstlisting}
print((4*7054)).
\end{lstlisting}
\begin{tabular}{ l l }
{\bf Target:} & 28216. \\
{\bf "Baseline" prediction:} & {\color[rgb]{0.31, 0.49, 0.16} 2}{\color[rgb]{0.31, 0.49, 0.16} 8}{\color[rgb]{0.31, 0.49, 0.16} 2}{\color[rgb]{0.31, 0.49, 0.16} 1}{\color[rgb]{0.31, 0.49, 0.16} 6}{\color[rgb]{0.31, 0.49, 0.16} .}\\
{\bf "Naive" prediction:} & {\color[rgb]{0.31, 0.49, 0.16} 2}{\color[rgb]{0.31, 0.49, 0.16} 8}{\color[rgb]{0.8,0.2,0.2} 1}{\color[rgb]{0.31, 0.49, 0.16} 1}{\color[rgb]{0.31, 0.49, 0.16} 6}{\color[rgb]{0.31, 0.49, 0.16} .}\\
{\bf "Mix" prediction:} & {\color[rgb]{0.31, 0.49, 0.16} 2}{\color[rgb]{0.31, 0.49, 0.16} 8}{\color[rgb]{0.31, 0.49, 0.16} 2}{\color[rgb]{0.31, 0.49, 0.16} 1}{\color[rgb]{0.31, 0.49, 0.16} 6}{\color[rgb]{0.31, 0.49, 0.16} .}\\
{\bf "Combined" prediction:} & {\color[rgb]{0.31, 0.49, 0.16} 2}{\color[rgb]{0.31, 0.49, 0.16} 8}{\color[rgb]{0.31, 0.49, 0.16} 2}{\color[rgb]{0.31, 0.49, 0.16} 1}{\color[rgb]{0.31, 0.49, 0.16} 6}{\color[rgb]{0.31, 0.49, 0.16} .}\\
\end{tabular}

\end{mdframed}\vspace{0.07cm}
\vspace{-0.1cm}\begin{mdframed}\vspace{0.07cm}
{\bf Input:}
\begin{lstlisting}
print((4635-5257)).
\end{lstlisting}
\begin{tabular}{ l l }
{\bf Target:} & -622. \\
{\bf "Baseline" prediction:} & {\color[rgb]{0.31, 0.49, 0.16} -}{\color[rgb]{0.31, 0.49, 0.16} 6}{\color[rgb]{0.8,0.2,0.2} 8}{\color[rgb]{0.8,0.2,0.2} 8}{\color[rgb]{0.31, 0.49, 0.16} .}\\
{\bf "Naive" prediction:} & {\color[rgb]{0.31, 0.49, 0.16} -}{\color[rgb]{0.31, 0.49, 0.16} 6}{\color[rgb]{0.31, 0.49, 0.16} 2}{\color[rgb]{0.8,0.2,0.2} 8}{\color[rgb]{0.31, 0.49, 0.16} .}\\
{\bf "Mix" prediction:} & {\color[rgb]{0.31, 0.49, 0.16} -}{\color[rgb]{0.31, 0.49, 0.16} 6}{\color[rgb]{0.8,0.2,0.2} 9}{\color[rgb]{0.31, 0.49, 0.16} 2}{\color[rgb]{0.31, 0.49, 0.16} .}\\
{\bf "Combined" prediction:} & {\color[rgb]{0.31, 0.49, 0.16} -}{\color[rgb]{0.31, 0.49, 0.16} 6}{\color[rgb]{0.8,0.2,0.2} 3}{\color[rgb]{0.31, 0.49, 0.16} 2}{\color[rgb]{0.31, 0.49, 0.16} .}\\
\end{tabular}

\end{mdframed}\vspace{0.07cm}
\vspace{-0.1cm}\begin{mdframed}\vspace{0.07cm}
{\bf Input:}
\begin{lstlisting}
e=1079
for x in range(10):e+=4729
print(e).
\end{lstlisting}
\begin{tabular}{ l l }
{\bf Target:} & 48369. \\
{\bf "Baseline" prediction:} & {\color[rgb]{0.31, 0.49, 0.16} 4}{\color[rgb]{0.31, 0.49, 0.16} 8}{\color[rgb]{0.8,0.2,0.2} 0}{\color[rgb]{0.8,0.2,0.2} 1}{\color[rgb]{0.8,0.2,0.2} 7}{\color[rgb]{0.31, 0.49, 0.16} .}\\
{\bf "Naive" prediction:} & {\color[rgb]{0.31, 0.49, 0.16} 4}{\color[rgb]{0.31, 0.49, 0.16} 8}{\color[rgb]{0.8,0.2,0.2} 0}{\color[rgb]{0.8,0.2,0.2} 1}{\color[rgb]{0.8,0.2,0.2} 1}{\color[rgb]{0.31, 0.49, 0.16} .}\\
{\bf "Mix" prediction:} & {\color[rgb]{0.31, 0.49, 0.16} 4}{\color[rgb]{0.31, 0.49, 0.16} 8}{\color[rgb]{0.8,0.2,0.2} 1}{\color[rgb]{0.8,0.2,0.2} 0}{\color[rgb]{0.8,0.2,0.2} 1}{\color[rgb]{0.31, 0.49, 0.16} .}\\
{\bf "Combined" prediction:} & {\color[rgb]{0.31, 0.49, 0.16} 4}{\color[rgb]{0.31, 0.49, 0.16} 8}{\color[rgb]{0.8,0.2,0.2} 0}{\color[rgb]{0.8,0.2,0.2} 0}{\color[rgb]{0.31, 0.49, 0.16} 9}{\color[rgb]{0.31, 0.49, 0.16} .}\\
\end{tabular}

\end{mdframed}\vspace{0.07cm}
\vspace{-0.1cm}

\vspace{0.3cm}\subsection{Examples of program evaluation prediction. Length = 4, Nesting = 2}\vspace{0.3cm}

\begin{mdframed}\vspace{0.07cm}
{\bf Input:}
\begin{lstlisting}
e=6653
for x in range(14):e+=6311
print(e).
\end{lstlisting}
\begin{tabular}{ l l }
{\bf Target:} & 95007. \\
{\bf "Baseline" prediction:} & {\color[rgb]{0.31, 0.49, 0.16} 9}{\color[rgb]{0.8,0.2,0.2} 4}{\color[rgb]{0.31, 0.49, 0.16} 0}{\color[rgb]{0.8,0.2,0.2} 9}{\color[rgb]{0.8,0.2,0.2} 3}{\color[rgb]{0.31, 0.49, 0.16} .}\\
{\bf "Naive" prediction:} & {\color[rgb]{0.31, 0.49, 0.16} 9}{\color[rgb]{0.8,0.2,0.2} 0}{\color[rgb]{0.31, 0.49, 0.16} 0}{\color[rgb]{0.8,0.2,0.2} 1}{\color[rgb]{0.8,0.2,0.2} 3}{\color[rgb]{0.31, 0.49, 0.16} .}\\
{\bf "Mix" prediction:} & {\color[rgb]{0.31, 0.49, 0.16} 9}{\color[rgb]{0.31, 0.49, 0.16} 5}{\color[rgb]{0.31, 0.49, 0.16} 0}{\color[rgb]{0.8,0.2,0.2} 1}{\color[rgb]{0.8,0.2,0.2} 5}{\color[rgb]{0.31, 0.49, 0.16} .}\\
{\bf "Combined" prediction:} & {\color[rgb]{0.31, 0.49, 0.16} 9}{\color[rgb]{0.8,0.2,0.2} 4}{\color[rgb]{0.8,0.2,0.2} 1}{\color[rgb]{0.31, 0.49, 0.16} 0}{\color[rgb]{0.8,0.2,0.2} 3}{\color[rgb]{0.31, 0.49, 0.16} .}\\
\end{tabular}

\end{mdframed}\vspace{0.07cm}
\vspace{-0.1cm}\begin{mdframed}\vspace{0.07cm}
{\bf Input:}
\begin{lstlisting}
i=6404;
print((i+8074)).
\end{lstlisting}
\begin{tabular}{ l l }
{\bf Target:} & 14478. \\
{\bf "Baseline" prediction:} & {\color[rgb]{0.31, 0.49, 0.16} 1}{\color[rgb]{0.31, 0.49, 0.16} 4}{\color[rgb]{0.31, 0.49, 0.16} 4}{\color[rgb]{0.8,0.2,0.2} 9}{\color[rgb]{0.31, 0.49, 0.16} 8}{\color[rgb]{0.31, 0.49, 0.16} .}\\
{\bf "Naive" prediction:} & {\color[rgb]{0.31, 0.49, 0.16} 1}{\color[rgb]{0.31, 0.49, 0.16} 4}{\color[rgb]{0.31, 0.49, 0.16} 4}{\color[rgb]{0.8,0.2,0.2} 4}{\color[rgb]{0.8,0.2,0.2} 4}{\color[rgb]{0.31, 0.49, 0.16} .}\\
{\bf "Mix" prediction:} & {\color[rgb]{0.31, 0.49, 0.16} 1}{\color[rgb]{0.31, 0.49, 0.16} 4}{\color[rgb]{0.31, 0.49, 0.16} 4}{\color[rgb]{0.8,0.2,0.2} 8}{\color[rgb]{0.8,0.2,0.2} 2}{\color[rgb]{0.31, 0.49, 0.16} .}\\
{\bf "Combined" prediction:} & {\color[rgb]{0.31, 0.49, 0.16} 1}{\color[rgb]{0.31, 0.49, 0.16} 4}{\color[rgb]{0.31, 0.49, 0.16} 4}{\color[rgb]{0.31, 0.49, 0.16} 7}{\color[rgb]{0.31, 0.49, 0.16} 8}{\color[rgb]{0.31, 0.49, 0.16} .}\\
\end{tabular}

\end{mdframed}\vspace{0.07cm}
\vspace{-0.1cm}\begin{mdframed}\vspace{0.07cm}
{\bf Input:}
\begin{lstlisting}
print((8*(5051-648))).
\end{lstlisting}
\begin{tabular}{ l l }
{\bf Target:} & 35224. \\
{\bf "Baseline" prediction:} & {\color[rgb]{0.31, 0.49, 0.16} 3}{\color[rgb]{0.8,0.2,0.2} 4}{\color[rgb]{0.8,0.2,0.2} 0}{\color[rgb]{0.8,0.2,0.2} 4}{\color[rgb]{0.31, 0.49, 0.16} 4}{\color[rgb]{0.31, 0.49, 0.16} .}\\
{\bf "Naive" prediction:} & {\color[rgb]{0.31, 0.49, 0.16} 3}{\color[rgb]{0.8,0.2,0.2} 2}{\color[rgb]{0.8,0.2,0.2} 1}{\color[rgb]{0.8,0.2,0.2} 8}{\color[rgb]{0.8,0.2,0.2} 0}{\color[rgb]{0.31, 0.49, 0.16} .}\\
{\bf "Mix" prediction:} & {\color[rgb]{0.31, 0.49, 0.16} 3}{\color[rgb]{0.8,0.2,0.2} 3}{\color[rgb]{0.31, 0.49, 0.16} 2}{\color[rgb]{0.8,0.2,0.2} 8}{\color[rgb]{0.31, 0.49, 0.16} 4}{\color[rgb]{0.31, 0.49, 0.16} .}\\
{\bf "Combined" prediction:} & {\color[rgb]{0.31, 0.49, 0.16} 3}{\color[rgb]{0.8,0.2,0.2} 3}{\color[rgb]{0.8,0.2,0.2} 0}{\color[rgb]{0.8,0.2,0.2} 0}{\color[rgb]{0.31, 0.49, 0.16} 4}{\color[rgb]{0.31, 0.49, 0.16} .}\\
\end{tabular}

\end{mdframed}\vspace{0.07cm}
\vspace{-0.1cm}\begin{mdframed}\vspace{0.07cm}
{\bf Input:}
\begin{lstlisting}
h=(3681 if 9279<3033 else 6191)
for x in range(7):h-=9910
print(h).
\end{lstlisting}
\begin{tabular}{ l l }
{\bf Target:} & -63179. \\
{\bf "Baseline" prediction:} & {\color[rgb]{0.31, 0.49, 0.16} -}{\color[rgb]{0.31, 0.49, 0.16} 6}{\color[rgb]{0.8,0.2,0.2} 2}{\color[rgb]{0.8,0.2,0.2} 0}{\color[rgb]{0.8,0.2,0.2} 4}{\color[rgb]{0.31, 0.49, 0.16} 9}{\color[rgb]{0.31, 0.49, 0.16} .}\\
{\bf "Naive" prediction:} & {\color[rgb]{0.31, 0.49, 0.16} -}{\color[rgb]{0.31, 0.49, 0.16} 6}{\color[rgb]{0.31, 0.49, 0.16} 3}{\color[rgb]{0.31, 0.49, 0.16} 1}{\color[rgb]{0.8,0.2,0.2} 1}{\color[rgb]{0.8,0.2,0.2} 7}{\color[rgb]{0.31, 0.49, 0.16} .}\\
{\bf "Mix" prediction:} & {\color[rgb]{0.31, 0.49, 0.16} -}{\color[rgb]{0.31, 0.49, 0.16} 6}{\color[rgb]{0.8,0.2,0.2} 2}{\color[rgb]{0.8,0.2,0.2} 0}{\color[rgb]{0.8,0.2,0.2} 1}{\color[rgb]{0.8,0.2,0.2} 3}{\color[rgb]{0.31, 0.49, 0.16} .}\\
{\bf "Combined" prediction:} & {\color[rgb]{0.31, 0.49, 0.16} -}{\color[rgb]{0.31, 0.49, 0.16} 6}{\color[rgb]{0.8,0.2,0.2} 2}{\color[rgb]{0.8,0.2,0.2} 0}{\color[rgb]{0.8,0.2,0.2} 0}{\color[rgb]{0.31, 0.49, 0.16} 9}{\color[rgb]{0.31, 0.49, 0.16} .}\\
\end{tabular}

\end{mdframed}\vspace{0.07cm}
\vspace{-0.1cm}\begin{mdframed}\vspace{0.07cm}
{\bf Input:}
\begin{lstlisting}
print(((3210+2472)+1477)).
\end{lstlisting}
\begin{tabular}{ l l }
{\bf Target:} & 7159. \\
{\bf "Baseline" prediction:} & {\color[rgb]{0.31, 0.49, 0.16} 7}{\color[rgb]{0.8,0.2,0.2} 0}{\color[rgb]{0.8,0.2,0.2} 0}{\color[rgb]{0.31, 0.49, 0.16} 9}{\color[rgb]{0.31, 0.49, 0.16} .}\\
{\bf "Naive" prediction:} & {\color[rgb]{0.31, 0.49, 0.16} 7}{\color[rgb]{0.8,0.2,0.2} 0}{\color[rgb]{0.8,0.2,0.2} 1}{\color[rgb]{0.31, 0.49, 0.16} 9}{\color[rgb]{0.31, 0.49, 0.16} .}\\
{\bf "Mix" prediction:} & {\color[rgb]{0.31, 0.49, 0.16} 7}{\color[rgb]{0.8,0.2,0.2} 9}{\color[rgb]{0.8,0.2,0.2} 9}{\color[rgb]{0.8,0.2,0.2} 5}{\color[rgb]{0.31, 0.49, 0.16} .}\\
{\bf "Combined" prediction:} & {\color[rgb]{0.31, 0.49, 0.16} 7}{\color[rgb]{0.8,0.2,0.2} 0}{\color[rgb]{0.8,0.2,0.2} 7}{\color[rgb]{0.31, 0.49, 0.16} 9}{\color[rgb]{0.31, 0.49, 0.16} .}\\
\end{tabular}

\end{mdframed}\vspace{0.07cm}
\vspace{-0.1cm}\begin{mdframed}\vspace{0.07cm}
{\bf Input:}
\begin{lstlisting}
b=8494
for x in range(2):b+=7484
print((b*14)).
\end{lstlisting}
\begin{tabular}{ l l }
{\bf Target:} & 328468. \\
{\bf "Baseline" prediction:} & {\color[rgb]{0.31, 0.49, 0.16} 3}{\color[rgb]{0.8,0.2,0.2} 1}{\color[rgb]{0.31, 0.49, 0.16} 8}{\color[rgb]{0.8,0.2,0.2} 0}{\color[rgb]{0.8,0.2,0.2} 0}{\color[rgb]{0.8,0.2,0.2} 4}{\color[rgb]{0.31, 0.49, 0.16} .}\\
{\bf "Naive" prediction:} & {\color[rgb]{0.31, 0.49, 0.16} 3}{\color[rgb]{0.8,0.2,0.2} 3}{\color[rgb]{0.31, 0.49, 0.16} 8}{\color[rgb]{0.8,0.2,0.2} 0}{\color[rgb]{0.8,0.2,0.2} 8}{\color[rgb]{0.31, 0.49, 0.16} 8}{\color[rgb]{0.31, 0.49, 0.16} .}\\
{\bf "Mix" prediction:} & {\color[rgb]{0.31, 0.49, 0.16} 3}{\color[rgb]{0.31, 0.49, 0.16} 2}{\color[rgb]{0.8,0.2,0.2} 9}{\color[rgb]{0.8,0.2,0.2} 2}{\color[rgb]{0.8,0.2,0.2} 2}{\color[rgb]{0.8,0.2,0.2} 0}{\color[rgb]{0.31, 0.49, 0.16} .}\\
{\bf "Combined" prediction:} & {\color[rgb]{0.31, 0.49, 0.16} 3}{\color[rgb]{0.8,0.2,0.2} 3}{\color[rgb]{0.31, 0.49, 0.16} 8}{\color[rgb]{0.8,0.2,0.2} 0}{\color[rgb]{0.8,0.2,0.2} 8}{\color[rgb]{0.8,0.2,0.2} 0}{\color[rgb]{0.31, 0.49, 0.16} .}\\
\end{tabular}

\end{mdframed}\vspace{0.07cm}
\vspace{-0.1cm}\begin{mdframed}\vspace{0.07cm}
{\bf Input:}
\begin{lstlisting}
j=6447;
print((12*(j-4689))).
\end{lstlisting}
\begin{tabular}{ l l }
{\bf Target:} & 21096. \\
{\bf "Baseline" prediction:} & {\color[rgb]{0.31, 0.49, 0.16} 2}{\color[rgb]{0.31, 0.49, 0.16} 1}{\color[rgb]{0.8,0.2,0.2} 2}{\color[rgb]{0.8,0.2,0.2} 6}{\color[rgb]{0.31, 0.49, 0.16} 6}{\color[rgb]{0.31, 0.49, 0.16} .}\\
{\bf "Naive" prediction:} & {\color[rgb]{0.8,0.2,0.2} 1}{\color[rgb]{0.8,0.2,0.2} 0}{\color[rgb]{0.31, 0.49, 0.16} 0}{\color[rgb]{0.8,0.2,0.2} 4}{\color[rgb]{0.31, 0.49, 0.16} 6}{\color[rgb]{0.31, 0.49, 0.16} .}\\
{\bf "Mix" prediction:} & {\color[rgb]{0.8,0.2,0.2} 1}{\color[rgb]{0.8,0.2,0.2} 0}{\color[rgb]{0.8,0.2,0.2} 6}{\color[rgb]{0.8,0.2,0.2} 0}{\color[rgb]{0.31, 0.49, 0.16} 6}{\color[rgb]{0.31, 0.49, 0.16} .}\\
{\bf "Combined" prediction:} & {\color[rgb]{0.31, 0.49, 0.16} 2}{\color[rgb]{0.8,0.2,0.2} 0}{\color[rgb]{0.8,0.2,0.2} 4}{\color[rgb]{0.8,0.2,0.2} 0}{\color[rgb]{0.8,0.2,0.2} 2}{\color[rgb]{0.31, 0.49, 0.16} .}\\
\end{tabular}

\end{mdframed}\vspace{0.07cm}
\vspace{-0.1cm}\begin{mdframed}\vspace{0.07cm}
{\bf Input:}
\begin{lstlisting}
print((13*9201)).
\end{lstlisting}
\begin{tabular}{ l l }
{\bf Target:} & 119613. \\
{\bf "Baseline" prediction:} & {\color[rgb]{0.31, 0.49, 0.16} 1}{\color[rgb]{0.31, 0.49, 0.16} 1}{\color[rgb]{0.8,0.2,0.2} 8}{\color[rgb]{0.8,0.2,0.2} 3}{\color[rgb]{0.31, 0.49, 0.16} 1}{\color[rgb]{0.31, 0.49, 0.16} 3}{\color[rgb]{0.31, 0.49, 0.16} .}\\
{\bf "Naive" prediction:} & {\color[rgb]{0.31, 0.49, 0.16} 1}{\color[rgb]{0.31, 0.49, 0.16} 1}{\color[rgb]{0.8,0.2,0.2} 8}{\color[rgb]{0.8,0.2,0.2} 0}{\color[rgb]{0.31, 0.49, 0.16} 1}{\color[rgb]{0.8,0.2,0.2} 1}{\color[rgb]{0.31, 0.49, 0.16} .}\\
{\bf "Mix" prediction:} & {\color[rgb]{0.31, 0.49, 0.16} 1}{\color[rgb]{0.31, 0.49, 0.16} 1}{\color[rgb]{0.8,0.2,0.2} 7}{\color[rgb]{0.31, 0.49, 0.16} 6}{\color[rgb]{0.8,0.2,0.2} 6}{\color[rgb]{0.8,0.2,0.2} 9}{\color[rgb]{0.31, 0.49, 0.16} .}\\
{\bf "Combined" prediction:} & {\color[rgb]{0.31, 0.49, 0.16} 1}{\color[rgb]{0.31, 0.49, 0.16} 1}{\color[rgb]{0.31, 0.49, 0.16} 9}{\color[rgb]{0.8,0.2,0.2} 5}{\color[rgb]{0.8,0.2,0.2} 3}{\color[rgb]{0.31, 0.49, 0.16} 3}{\color[rgb]{0.31, 0.49, 0.16} .}\\
\end{tabular}

\end{mdframed}\vspace{0.07cm}
\vspace{-0.1cm}\begin{mdframed}\vspace{0.07cm}
{\bf Input:}
\begin{lstlisting}
g=1054;
print((6028+(g-1953))).
\end{lstlisting}
\begin{tabular}{ l l }
{\bf Target:} & 5129. \\
{\bf "Baseline" prediction:} & {\color[rgb]{0.8,0.2,0.2} 4}{\color[rgb]{0.8,0.2,0.2} 0}{\color[rgb]{0.8,0.2,0.2} 1}{\color[rgb]{0.8,0.2,0.2} 3}{\color[rgb]{0.31, 0.49, 0.16} .}\\
{\bf "Naive" prediction:} & {\color[rgb]{0.31, 0.49, 0.16} 5}{\color[rgb]{0.8,0.2,0.2} 0}{\color[rgb]{0.8,0.2,0.2} 3}{\color[rgb]{0.8,0.2,0.2} 5}{\color[rgb]{0.31, 0.49, 0.16} .}\\
{\bf "Mix" prediction:} & {\color[rgb]{0.8,0.2,0.2} 4}{\color[rgb]{0.8,0.2,0.2} 0}{\color[rgb]{0.8,0.2,0.2} 1}{\color[rgb]{0.8,0.2,0.2} 5}{\color[rgb]{0.31, 0.49, 0.16} .}\\
{\bf "Combined" prediction:} & {\color[rgb]{0.8,0.2,0.2} 4}{\color[rgb]{0.8,0.2,0.2} 0}{\color[rgb]{0.8,0.2,0.2} 0}{\color[rgb]{0.31, 0.49, 0.16} 9}{\color[rgb]{0.31, 0.49, 0.16} .}\\
\end{tabular}

\end{mdframed}\vspace{0.07cm}
\vspace{-0.1cm}\begin{mdframed}\vspace{0.07cm}
{\bf Input:}
\begin{lstlisting}
d=6817
for x in range(7):d-=(4581-2186)
print(d).
\end{lstlisting}
\begin{tabular}{ l l }
{\bf Target:} & -9948. \\
{\bf "Baseline" prediction:} & {\color[rgb]{0.31, 0.49, 0.16} -}{\color[rgb]{0.8,0.2,0.2} 1}{\color[rgb]{0.31, 0.49, 0.16} 9}{\color[rgb]{0.8,0.2,0.2} 9}{\color[rgb]{0.8,0.2,0.2} 6}{\color[rgb]{0.31, 0.49, 0.16} .}\\
{\bf "Naive" prediction:} & {\color[rgb]{0.31, 0.49, 0.16} -}{\color[rgb]{0.8,0.2,0.2} 1}{\color[rgb]{0.8,0.2,0.2} 6}{\color[rgb]{0.8,0.2,0.2} 1}{\color[rgb]{0.8,0.2,0.2} 0}{\color[rgb]{0.31, 0.49, 0.16} .}\\
{\bf "Mix" prediction:} & {\color[rgb]{0.31, 0.49, 0.16} -}{\color[rgb]{0.8,0.2,0.2} 1}{\color[rgb]{0.8,0.2,0.2} 8}{\color[rgb]{0.8,0.2,0.2} 8}{\color[rgb]{0.8,0.2,0.2} 2}{\color[rgb]{0.31, 0.49, 0.16} .}\\
{\bf "Combined" prediction:} & {\color[rgb]{0.31, 0.49, 0.16} -}{\color[rgb]{0.8,0.2,0.2} 1}{\color[rgb]{0.31, 0.49, 0.16} 9}{\color[rgb]{0.8,0.2,0.2} 8}{\color[rgb]{0.8,0.2,0.2} 0}{\color[rgb]{0.31, 0.49, 0.16} .}\\
\end{tabular}

\end{mdframed}\vspace{0.07cm}
\vspace{-0.1cm}

\vspace{0.3cm}\subsection{Examples of program evaluation prediction. Length = 4, Nesting = 3}\vspace{0.3cm}

\begin{mdframed}\vspace{0.07cm}
{\bf Input:}
\begin{lstlisting}
f=4692
for x in range(4):f-=1664
j=1443
for x in range(8):j+=f
d=j
for x in range(11):d-=4699
print(d).
\end{lstlisting}
\begin{tabular}{ l l }
{\bf Target:} & -65958. \\
{\bf "Baseline" prediction:} & {\color[rgb]{0.31, 0.49, 0.16} -}{\color[rgb]{0.8,0.2,0.2} 1}{\color[rgb]{0.8,0.2,0.2} 3}{\color[rgb]{0.8,0.2,0.2} 2}{\color[rgb]{0.8,0.2,0.2} 6}{\color[rgb]{0.8,0.2,0.2} 2}{\color[rgb]{0.31, 0.49, 0.16} .}\\
{\bf "Naive" prediction:} & {\color[rgb]{0.31, 0.49, 0.16} -}{\color[rgb]{0.8,0.2,0.2} 7}{\color[rgb]{0.8,0.2,0.2} 3}{\color[rgb]{0.8,0.2,0.2} 1}{\color[rgb]{0.8,0.2,0.2} 9}{\color[rgb]{0.8,0.2,0.2} 4}{\color[rgb]{0.31, 0.49, 0.16} .}\\
{\bf "Mix" prediction:} & {\color[rgb]{0.31, 0.49, 0.16} -}{\color[rgb]{0.8,0.2,0.2} 4}{\color[rgb]{0.8,0.2,0.2} 0}{\color[rgb]{0.8,0.2,0.2} 1}{\color[rgb]{0.8,0.2,0.2} 8}{\color[rgb]{0.31, 0.49, 0.16} 8}{\color[rgb]{0.31, 0.49, 0.16} .}\\
{\bf "Combined" prediction:} & {\color[rgb]{0.31, 0.49, 0.16} -}{\color[rgb]{0.8,0.2,0.2} 1}{\color[rgb]{0.8,0.2,0.2} 2}{\color[rgb]{0.8,0.2,0.2} 0}{\color[rgb]{0.8,0.2,0.2} 0}{\color[rgb]{0.8,0.2,0.2} 4}{\color[rgb]{0.31, 0.49, 0.16} .}\\
\end{tabular}

\end{mdframed}\vspace{0.07cm}
\vspace{-0.1cm}\begin{mdframed}\vspace{0.07cm}
{\bf Input:}
\begin{lstlisting}
b=9930
for x in range(11):b-=4369
g=b;
print(((g-8043)+9955)).
\end{lstlisting}
\begin{tabular}{ l l }
{\bf Target:} & -36217. \\
{\bf "Baseline" prediction:} & {\color[rgb]{0.31, 0.49, 0.16} -}{\color[rgb]{0.31, 0.49, 0.16} 3}{\color[rgb]{0.8,0.2,0.2} 7}{\color[rgb]{0.8,0.2,0.2} 5}{\color[rgb]{0.31, 0.49, 0.16} 1}{\color[rgb]{0.8,0.2,0.2} 5}{\color[rgb]{0.31, 0.49, 0.16} .}\\
{\bf "Naive" prediction:} & {\color[rgb]{0.31, 0.49, 0.16} -}{\color[rgb]{0.31, 0.49, 0.16} 3}{\color[rgb]{0.8,0.2,0.2} 8}{\color[rgb]{0.8,0.2,0.2} 6}{\color[rgb]{0.8,0.2,0.2} 0}{\color[rgb]{0.8,0.2,0.2} 9}{\color[rgb]{0.31, 0.49, 0.16} .}\\
{\bf "Mix" prediction:} & {\color[rgb]{0.31, 0.49, 0.16} -}{\color[rgb]{0.31, 0.49, 0.16} 3}{\color[rgb]{0.8,0.2,0.2} 5}{\color[rgb]{0.8,0.2,0.2} 8}{\color[rgb]{0.8,0.2,0.2} 9}{\color[rgb]{0.8,0.2,0.2} 3}{\color[rgb]{0.31, 0.49, 0.16} .}\\
{\bf "Combined" prediction:} & {\color[rgb]{0.31, 0.49, 0.16} -}{\color[rgb]{0.31, 0.49, 0.16} 3}{\color[rgb]{0.8,0.2,0.2} 5}{\color[rgb]{0.8,0.2,0.2} 0}{\color[rgb]{0.8,0.2,0.2} 5}{\color[rgb]{0.8,0.2,0.2} 5}{\color[rgb]{0.31, 0.49, 0.16} .}\\
\end{tabular}

\end{mdframed}\vspace{0.07cm}
\vspace{-0.1cm}\begin{mdframed}\vspace{0.07cm}
{\bf Input:}
\begin{lstlisting}
d=5446
for x in range(8):d+=(2678 if 4803<2829 else 9848)
print((d if 5935<4845 else 3043)).
\end{lstlisting}
\begin{tabular}{ l l }
{\bf Target:} & 3043. \\
{\bf "Baseline" prediction:} & {\color[rgb]{0.31, 0.49, 0.16} 3}{\color[rgb]{0.31, 0.49, 0.16} 0}{\color[rgb]{0.31, 0.49, 0.16} 4}{\color[rgb]{0.31, 0.49, 0.16} 3}{\color[rgb]{0.31, 0.49, 0.16} .}\\
{\bf "Naive" prediction:} & {\color[rgb]{0.31, 0.49, 0.16} 3}{\color[rgb]{0.31, 0.49, 0.16} 0}{\color[rgb]{0.31, 0.49, 0.16} 4}{\color[rgb]{0.31, 0.49, 0.16} 3}{\color[rgb]{0.31, 0.49, 0.16} .}\\
{\bf "Mix" prediction:} & {\color[rgb]{0.31, 0.49, 0.16} 3}{\color[rgb]{0.31, 0.49, 0.16} 0}{\color[rgb]{0.31, 0.49, 0.16} 4}{\color[rgb]{0.31, 0.49, 0.16} 3}{\color[rgb]{0.31, 0.49, 0.16} .}\\
{\bf "Combined" prediction:} & {\color[rgb]{0.31, 0.49, 0.16} 3}{\color[rgb]{0.31, 0.49, 0.16} 0}{\color[rgb]{0.31, 0.49, 0.16} 4}{\color[rgb]{0.31, 0.49, 0.16} 3}{\color[rgb]{0.31, 0.49, 0.16} .}\\
\end{tabular}

\end{mdframed}\vspace{0.07cm}
\vspace{-0.1cm}\begin{mdframed}\vspace{0.07cm}
{\bf Input:}
\begin{lstlisting}
print((((2578 if 7750<1768 else 8639)-2590)+342)).
\end{lstlisting}
\begin{tabular}{ l l }
{\bf Target:} & 6391. \\
{\bf "Baseline" prediction:} & {\color[rgb]{0.8,0.2,0.2} -}{\color[rgb]{0.8,0.2,0.2} 5}{\color[rgb]{0.8,0.2,0.2} 5}{\color[rgb]{0.8,0.2,0.2} 5}{\color[rgb]{0.31, 0.49, 0.16} .}\\
{\bf "Naive" prediction:} & {\color[rgb]{0.31, 0.49, 0.16} 6}{\color[rgb]{0.31, 0.49, 0.16} 3}{\color[rgb]{0.8,0.2,0.2} 2}{\color[rgb]{0.8,0.2,0.2} 9}{\color[rgb]{0.31, 0.49, 0.16} .}\\
{\bf "Mix" prediction:} & {\color[rgb]{0.31, 0.49, 0.16} 6}{\color[rgb]{0.8,0.2,0.2} 4}{\color[rgb]{0.8,0.2,0.2} 6}{\color[rgb]{0.31, 0.49, 0.16} 1}{\color[rgb]{0.31, 0.49, 0.16} .}\\
{\bf "Combined" prediction:} & {\color[rgb]{0.31, 0.49, 0.16} 6}{\color[rgb]{0.8,0.2,0.2} 1}{\color[rgb]{0.8,0.2,0.2} 0}{\color[rgb]{0.8,0.2,0.2} 5}{\color[rgb]{0.31, 0.49, 0.16} .}\\
\end{tabular}

\end{mdframed}\vspace{0.07cm}
\vspace{-0.1cm}\begin{mdframed}\vspace{0.07cm}
{\bf Input:}
\begin{lstlisting}
print((((841 if 2076<7326 else 1869)*10) if 7827<317 else 7192)).
\end{lstlisting}
\begin{tabular}{ l l }
{\bf Target:} & 7192. \\
{\bf "Baseline" prediction:} & {\color[rgb]{0.31, 0.49, 0.16} 7}{\color[rgb]{0.31, 0.49, 0.16} 1}{\color[rgb]{0.31, 0.49, 0.16} 9}{\color[rgb]{0.31, 0.49, 0.16} 2}{\color[rgb]{0.31, 0.49, 0.16} .}\\
{\bf "Naive" prediction:} & {\color[rgb]{0.31, 0.49, 0.16} 7}{\color[rgb]{0.31, 0.49, 0.16} 1}{\color[rgb]{0.31, 0.49, 0.16} 9}{\color[rgb]{0.31, 0.49, 0.16} 2}{\color[rgb]{0.31, 0.49, 0.16} .}\\
{\bf "Mix" prediction:} & {\color[rgb]{0.31, 0.49, 0.16} 7}{\color[rgb]{0.31, 0.49, 0.16} 1}{\color[rgb]{0.31, 0.49, 0.16} 9}{\color[rgb]{0.31, 0.49, 0.16} 2}{\color[rgb]{0.31, 0.49, 0.16} .}\\
{\bf "Combined" prediction:} & {\color[rgb]{0.31, 0.49, 0.16} 7}{\color[rgb]{0.31, 0.49, 0.16} 1}{\color[rgb]{0.31, 0.49, 0.16} 9}{\color[rgb]{0.31, 0.49, 0.16} 2}{\color[rgb]{0.31, 0.49, 0.16} .}\\
\end{tabular}

\end{mdframed}\vspace{0.07cm}
\vspace{-0.1cm}\begin{mdframed}\vspace{0.07cm}
{\bf Input:}
\begin{lstlisting}
d=8640;
print((7135 if 6710>((d+7080)*14) else 7200)).
\end{lstlisting}
\begin{tabular}{ l l }
{\bf Target:} & 7200. \\
{\bf "Baseline" prediction:} & {\color[rgb]{0.31, 0.49, 0.16} 7}{\color[rgb]{0.31, 0.49, 0.16} 2}{\color[rgb]{0.31, 0.49, 0.16} 0}{\color[rgb]{0.31, 0.49, 0.16} 0}{\color[rgb]{0.31, 0.49, 0.16} .}\\
{\bf "Naive" prediction:} & {\color[rgb]{0.31, 0.49, 0.16} 7}{\color[rgb]{0.31, 0.49, 0.16} 2}{\color[rgb]{0.31, 0.49, 0.16} 0}{\color[rgb]{0.31, 0.49, 0.16} 0}{\color[rgb]{0.31, 0.49, 0.16} .}\\
{\bf "Mix" prediction:} & {\color[rgb]{0.31, 0.49, 0.16} 7}{\color[rgb]{0.31, 0.49, 0.16} 2}{\color[rgb]{0.31, 0.49, 0.16} 0}{\color[rgb]{0.31, 0.49, 0.16} 0}{\color[rgb]{0.31, 0.49, 0.16} .}\\
{\bf "Combined" prediction:} & {\color[rgb]{0.31, 0.49, 0.16} 7}{\color[rgb]{0.31, 0.49, 0.16} 2}{\color[rgb]{0.31, 0.49, 0.16} 0}{\color[rgb]{0.31, 0.49, 0.16} 0}{\color[rgb]{0.31, 0.49, 0.16} .}\\
\end{tabular}

\end{mdframed}\vspace{0.07cm}
\vspace{-0.1cm}\begin{mdframed}\vspace{0.07cm}
{\bf Input:}
\begin{lstlisting}
b=6968
for x in range(10):b-=(299 if 3389<9977 else 203)
print((12*b)).
\end{lstlisting}
\begin{tabular}{ l l }
{\bf Target:} & 47736. \\
{\bf "Baseline" prediction:} & {\color[rgb]{0.8,0.2,0.2} -}{\color[rgb]{0.8,0.2,0.2} 0}{\color[rgb]{0.8,0.2,0.2} 6}{\color[rgb]{0.8,0.2,0.2} 6}{\color[rgb]{0.31, 0.49, 0.16} 6}{\color[rgb]{0.31, 0.49, 0.16} .}\\
{\bf "Naive" prediction:} & {\color[rgb]{0.8,0.2,0.2} 1}{\color[rgb]{0.8,0.2,0.2} 1}{\color[rgb]{0.8,0.2,0.2} 2}{\color[rgb]{0.8,0.2,0.2} 6}{\color[rgb]{0.8,0.2,0.2} 2}{\color[rgb]{0.31, 0.49, 0.16} .}\\
{\bf "Mix" prediction:} & {\color[rgb]{0.31, 0.49, 0.16} 4}{\color[rgb]{0.8,0.2,0.2} 8}{\color[rgb]{0.8,0.2,0.2} 6}{\color[rgb]{0.8,0.2,0.2} 6}{\color[rgb]{0.31, 0.49, 0.16} 6}{\color[rgb]{0.31, 0.49, 0.16} .}\\
{\bf "Combined" prediction:} & {\color[rgb]{0.31, 0.49, 0.16} 4}{\color[rgb]{0.8,0.2,0.2} 8}{\color[rgb]{0.31, 0.49, 0.16} 7}{\color[rgb]{0.8,0.2,0.2} 6}{\color[rgb]{0.31, 0.49, 0.16} 6}{\color[rgb]{0.31, 0.49, 0.16} .}\\
\end{tabular}

\end{mdframed}\vspace{0.07cm}
\vspace{-0.1cm}\begin{mdframed}\vspace{0.07cm}
{\bf Input:}
\begin{lstlisting}
j=(1*5057);
print(((j+1215)+6931)).
\end{lstlisting}
\begin{tabular}{ l l }
{\bf Target:} & 13203. \\
{\bf "Baseline" prediction:} & {\color[rgb]{0.31, 0.49, 0.16} 1}{\color[rgb]{0.31, 0.49, 0.16} 3}{\color[rgb]{0.8,0.2,0.2} 0}{\color[rgb]{0.8,0.2,0.2} 1}{\color[rgb]{0.8,0.2,0.2} 5}{\color[rgb]{0.31, 0.49, 0.16} .}\\
{\bf "Naive" prediction:} & {\color[rgb]{0.31, 0.49, 0.16} 1}{\color[rgb]{0.8,0.2,0.2} 2}{\color[rgb]{0.8,0.2,0.2} 0}{\color[rgb]{0.31, 0.49, 0.16} 0}{\color[rgb]{0.8,0.2,0.2} 7}{\color[rgb]{0.31, 0.49, 0.16} .}\\
{\bf "Mix" prediction:} & {\color[rgb]{0.31, 0.49, 0.16} 1}{\color[rgb]{0.31, 0.49, 0.16} 3}{\color[rgb]{0.8,0.2,0.2} 3}{\color[rgb]{0.8,0.2,0.2} 7}{\color[rgb]{0.8,0.2,0.2} 9}{\color[rgb]{0.31, 0.49, 0.16} .}\\
{\bf "Combined" prediction:} & {\color[rgb]{0.31, 0.49, 0.16} 1}{\color[rgb]{0.31, 0.49, 0.16} 3}{\color[rgb]{0.31, 0.49, 0.16} 2}{\color[rgb]{0.31, 0.49, 0.16} 0}{\color[rgb]{0.8,0.2,0.2} 5}{\color[rgb]{0.31, 0.49, 0.16} .}\\
\end{tabular}

\end{mdframed}\vspace{0.07cm}
\vspace{-0.1cm}\begin{mdframed}\vspace{0.07cm}
{\bf Input:}
\begin{lstlisting}
print(((1090-3305)+9466)).
\end{lstlisting}
\begin{tabular}{ l l }
{\bf Target:} & 7251. \\
{\bf "Baseline" prediction:} & {\color[rgb]{0.31, 0.49, 0.16} 7}{\color[rgb]{0.8,0.2,0.2} 1}{\color[rgb]{0.8,0.2,0.2} 1}{\color[rgb]{0.31, 0.49, 0.16} 1}{\color[rgb]{0.31, 0.49, 0.16} .}\\
{\bf "Naive" prediction:} & {\color[rgb]{0.31, 0.49, 0.16} 7}{\color[rgb]{0.8,0.2,0.2} 0}{\color[rgb]{0.8,0.2,0.2} 9}{\color[rgb]{0.8,0.2,0.2} 9}{\color[rgb]{0.31, 0.49, 0.16} .}\\
{\bf "Mix" prediction:} & {\color[rgb]{0.31, 0.49, 0.16} 7}{\color[rgb]{0.8,0.2,0.2} 5}{\color[rgb]{0.8,0.2,0.2} 9}{\color[rgb]{0.8,0.2,0.2} 5}{\color[rgb]{0.31, 0.49, 0.16} .}\\
{\bf "Combined" prediction:} & {\color[rgb]{0.31, 0.49, 0.16} 7}{\color[rgb]{0.8,0.2,0.2} 6}{\color[rgb]{0.8,0.2,0.2} 9}{\color[rgb]{0.8,0.2,0.2} 9}{\color[rgb]{0.31, 0.49, 0.16} .}\\
\end{tabular}

\end{mdframed}\vspace{0.07cm}
\vspace{-0.1cm}\begin{mdframed}\vspace{0.07cm}
{\bf Input:}
\begin{lstlisting}
a=8331;
print((a-(15*7082))).
\end{lstlisting}
\begin{tabular}{ l l }
{\bf Target:} & -97899. \\
{\bf "Baseline" prediction:} & {\color[rgb]{0.31, 0.49, 0.16} -}{\color[rgb]{0.31, 0.49, 0.16} 9}{\color[rgb]{0.8,0.2,0.2} 6}{\color[rgb]{0.8,0.2,0.2} 9}{\color[rgb]{0.31, 0.49, 0.16} 9}{\color[rgb]{0.8,0.2,0.2} 1}{\color[rgb]{0.31, 0.49, 0.16} .}\\
{\bf "Naive" prediction:} & {\color[rgb]{0.31, 0.49, 0.16} -}{\color[rgb]{0.8,0.2,0.2} 1}{\color[rgb]{0.8,0.2,0.2} 9}{\color[rgb]{0.8,0.2,0.2} 9}{\color[rgb]{0.8,0.2,0.2} 5}{\color[rgb]{0.31, 0.49, 0.16} 9}{\color[rgb]{0.31, 0.49, 0.16} .}\\
{\bf "Mix" prediction:} & {\color[rgb]{0.31, 0.49, 0.16} -}{\color[rgb]{0.31, 0.49, 0.16} 9}{\color[rgb]{0.8,0.2,0.2} 5}{\color[rgb]{0.8,0.2,0.2} 5}{\color[rgb]{0.8,0.2,0.2} 5}{\color[rgb]{0.8,0.2,0.2} 1}{\color[rgb]{0.31, 0.49, 0.16} .}\\
{\bf "Combined" prediction:} & {\color[rgb]{0.31, 0.49, 0.16} -}{\color[rgb]{0.31, 0.49, 0.16} 9}{\color[rgb]{0.8,0.2,0.2} 6}{\color[rgb]{0.8,0.2,0.2} 3}{\color[rgb]{0.31, 0.49, 0.16} 9}{\color[rgb]{0.8,0.2,0.2} 7}{\color[rgb]{0.31, 0.49, 0.16} .}\\
\end{tabular}

\end{mdframed}\vspace{0.07cm}
\vspace{-0.1cm}

\vspace{0.3cm}\subsection{Examples of program evaluation prediction. Length = 6, Nesting = 1}\vspace{0.3cm}

\begin{mdframed}\vspace{0.07cm}
{\bf Input:}
\begin{lstlisting}
print((71647-548966)).
\end{lstlisting}
\begin{tabular}{ l l }
{\bf Target:} & -477319. \\
{\bf "Baseline" prediction:} & {\color[rgb]{0.31, 0.49, 0.16} -}{\color[rgb]{0.31, 0.49, 0.16} 4}{\color[rgb]{0.31, 0.49, 0.16} 7}{\color[rgb]{0.8,0.2,0.2} 2}{\color[rgb]{0.8,0.2,0.2} 1}{\color[rgb]{0.8,0.2,0.2} 2}{\color[rgb]{0.8,0.2,0.2} 2}{\color[rgb]{0.31, 0.49, 0.16} .}\\
{\bf "Naive" prediction:} & {\color[rgb]{0.31, 0.49, 0.16} -}{\color[rgb]{0.31, 0.49, 0.16} 4}{\color[rgb]{0.31, 0.49, 0.16} 7}{\color[rgb]{0.31, 0.49, 0.16} 7}{\color[rgb]{0.8,0.2,0.2} 5}{\color[rgb]{0.8,0.2,0.2} 9}{\color[rgb]{0.8,0.2,0.2} 1}{\color[rgb]{0.31, 0.49, 0.16} .}\\
{\bf "Mix" prediction:} & {\color[rgb]{0.31, 0.49, 0.16} -}{\color[rgb]{0.31, 0.49, 0.16} 4}{\color[rgb]{0.31, 0.49, 0.16} 7}{\color[rgb]{0.8,0.2,0.2} 9}{\color[rgb]{0.8,0.2,0.2} 7}{\color[rgb]{0.8,0.2,0.2} 0}{\color[rgb]{0.8,0.2,0.2} 5}{\color[rgb]{0.31, 0.49, 0.16} .}\\
{\bf "Combined" prediction:} & {\color[rgb]{0.31, 0.49, 0.16} -}{\color[rgb]{0.31, 0.49, 0.16} 4}{\color[rgb]{0.31, 0.49, 0.16} 7}{\color[rgb]{0.8,0.2,0.2} 5}{\color[rgb]{0.8,0.2,0.2} 0}{\color[rgb]{0.8,0.2,0.2} 0}{\color[rgb]{0.31, 0.49, 0.16} 9}{\color[rgb]{0.31, 0.49, 0.16} .}\\
\end{tabular}

\end{mdframed}\vspace{0.07cm}
\vspace{-0.1cm}\begin{mdframed}\vspace{0.07cm}
{\bf Input:}
\begin{lstlisting}
print(1508).
\end{lstlisting}
\begin{tabular}{ l l }
{\bf Target:} & 1508. \\
{\bf "Baseline" prediction:} & {\color[rgb]{0.31, 0.49, 0.16} 1}{\color[rgb]{0.31, 0.49, 0.16} 5}{\color[rgb]{0.31, 0.49, 0.16} 0}{\color[rgb]{0.31, 0.49, 0.16} 8}{\color[rgb]{0.31, 0.49, 0.16} .}\\
{\bf "Naive" prediction:} & {\color[rgb]{0.31, 0.49, 0.16} 1}{\color[rgb]{0.31, 0.49, 0.16} 5}{\color[rgb]{0.31, 0.49, 0.16} 0}{\color[rgb]{0.31, 0.49, 0.16} 8}{\color[rgb]{0.31, 0.49, 0.16} .}\\
{\bf "Mix" prediction:} & {\color[rgb]{0.31, 0.49, 0.16} 1}{\color[rgb]{0.31, 0.49, 0.16} 5}{\color[rgb]{0.31, 0.49, 0.16} 0}{\color[rgb]{0.31, 0.49, 0.16} 8}{\color[rgb]{0.31, 0.49, 0.16} .}\\
{\bf "Combined" prediction:} & {\color[rgb]{0.31, 0.49, 0.16} 1}{\color[rgb]{0.31, 0.49, 0.16} 5}{\color[rgb]{0.31, 0.49, 0.16} 0}{\color[rgb]{0.31, 0.49, 0.16} 8}{\color[rgb]{0.31, 0.49, 0.16} .}\\
\end{tabular}

\end{mdframed}\vspace{0.07cm}
\vspace{-0.1cm}\begin{mdframed}\vspace{0.07cm}
{\bf Input:}
\begin{lstlisting}
j=611989;
print((j+763864)).
\end{lstlisting}
\begin{tabular}{ l l }
{\bf Target:} & 1375853. \\
{\bf "Baseline" prediction:} & {\color[rgb]{0.31, 0.49, 0.16} 1}{\color[rgb]{0.31, 0.49, 0.16} 3}{\color[rgb]{0.31, 0.49, 0.16} 7}{\color[rgb]{0.8,0.2,0.2} 9}{\color[rgb]{0.8,0.2,0.2} 9}{\color[rgb]{0.8,0.2,0.2} 2}{\color[rgb]{0.8,0.2,0.2} 0}{\color[rgb]{0.31, 0.49, 0.16} .}\\
{\bf "Naive" prediction:} & {\color[rgb]{0.31, 0.49, 0.16} 1}{\color[rgb]{0.31, 0.49, 0.16} 3}{\color[rgb]{0.31, 0.49, 0.16} 7}{\color[rgb]{0.8,0.2,0.2} 8}{\color[rgb]{0.8,0.2,0.2} 9}{\color[rgb]{0.8,0.2,0.2} 9}{\color[rgb]{0.8,0.2,0.2} 1}{\color[rgb]{0.31, 0.49, 0.16} .}\\
{\bf "Mix" prediction:} & {\color[rgb]{0.31, 0.49, 0.16} 1}{\color[rgb]{0.31, 0.49, 0.16} 3}{\color[rgb]{0.31, 0.49, 0.16} 7}{\color[rgb]{0.31, 0.49, 0.16} 5}{\color[rgb]{0.8,0.2,0.2} 1}{\color[rgb]{0.8,0.2,0.2} 1}{\color[rgb]{0.8,0.2,0.2} 9}{\color[rgb]{0.31, 0.49, 0.16} .}\\
{\bf "Combined" prediction:} & {\color[rgb]{0.31, 0.49, 0.16} 1}{\color[rgb]{0.31, 0.49, 0.16} 3}{\color[rgb]{0.31, 0.49, 0.16} 7}{\color[rgb]{0.31, 0.49, 0.16} 5}{\color[rgb]{0.8,0.2,0.2} 1}{\color[rgb]{0.8,0.2,0.2} 7}{\color[rgb]{0.31, 0.49, 0.16} 3}{\color[rgb]{0.31, 0.49, 0.16} .}\\
\end{tabular}

\end{mdframed}\vspace{0.07cm}
\vspace{-0.1cm}\begin{mdframed}\vspace{0.07cm}
{\bf Input:}
\begin{lstlisting}
print((151108 if 289653>33296 else 564130)).
\end{lstlisting}
\begin{tabular}{ l l }
{\bf Target:} & 151108. \\
{\bf "Baseline" prediction:} & {\color[rgb]{0.31, 0.49, 0.16} 1}{\color[rgb]{0.31, 0.49, 0.16} 5}{\color[rgb]{0.8,0.2,0.2} 4}{\color[rgb]{0.8,0.2,0.2} 9}{\color[rgb]{0.8,0.2,0.2} 7}{\color[rgb]{0.8,0.2,0.2} 3}{\color[rgb]{0.31, 0.49, 0.16} .}\\
{\bf "Naive" prediction:} & {\color[rgb]{0.31, 0.49, 0.16} 1}{\color[rgb]{0.31, 0.49, 0.16} 5}{\color[rgb]{0.31, 0.49, 0.16} 1}{\color[rgb]{0.31, 0.49, 0.16} 1}{\color[rgb]{0.31, 0.49, 0.16} 0}{\color[rgb]{0.31, 0.49, 0.16} 8}{\color[rgb]{0.31, 0.49, 0.16} .}\\
{\bf "Mix" prediction:} & {\color[rgb]{0.31, 0.49, 0.16} 1}{\color[rgb]{0.31, 0.49, 0.16} 5}{\color[rgb]{0.31, 0.49, 0.16} 1}{\color[rgb]{0.31, 0.49, 0.16} 1}{\color[rgb]{0.31, 0.49, 0.16} 0}{\color[rgb]{0.31, 0.49, 0.16} 8}{\color[rgb]{0.31, 0.49, 0.16} .}\\
{\bf "Combined" prediction:} & {\color[rgb]{0.31, 0.49, 0.16} 1}{\color[rgb]{0.31, 0.49, 0.16} 5}{\color[rgb]{0.31, 0.49, 0.16} 1}{\color[rgb]{0.31, 0.49, 0.16} 1}{\color[rgb]{0.31, 0.49, 0.16} 0}{\color[rgb]{0.31, 0.49, 0.16} 8}{\color[rgb]{0.31, 0.49, 0.16} .}\\
\end{tabular}

\end{mdframed}\vspace{0.07cm}
\vspace{-0.1cm}\begin{mdframed}\vspace{0.07cm}
{\bf Input:}
\begin{lstlisting}
c=142012
for x in range(12):c-=166776
print(c).
\end{lstlisting}
\begin{tabular}{ l l }
{\bf Target:} & -1859300. \\
{\bf "Baseline" prediction:} & {\color[rgb]{0.31, 0.49, 0.16} -}{\color[rgb]{0.31, 0.49, 0.16} 1}{\color[rgb]{0.31, 0.49, 0.16} 8}{\color[rgb]{0.8,0.2,0.2} 4}{\color[rgb]{0.8,0.2,0.2} 0}{\color[rgb]{0.8,0.2,0.2} 8}{\color[rgb]{0.8,0.2,0.2} 3}{\color[rgb]{0.8,0.2,0.2} 1}{\color[rgb]{0.31, 0.49, 0.16} .}\\
{\bf "Naive" prediction:} & {\color[rgb]{0.31, 0.49, 0.16} -}{\color[rgb]{0.31, 0.49, 0.16} 1}{\color[rgb]{0.31, 0.49, 0.16} 8}{\color[rgb]{0.8,0.2,0.2} 4}{\color[rgb]{0.8,0.2,0.2} 0}{\color[rgb]{0.8,0.2,0.2} 0}{\color[rgb]{0.31, 0.49, 0.16} 0}{\color[rgb]{0.31, 0.49, 0.16} 0}{\color[rgb]{0.31, 0.49, 0.16} .}\\
{\bf "Mix" prediction:} & {\color[rgb]{0.31, 0.49, 0.16} -}{\color[rgb]{0.31, 0.49, 0.16} 1}{\color[rgb]{0.8,0.2,0.2} 9}{\color[rgb]{0.8,0.2,0.2} 7}{\color[rgb]{0.31, 0.49, 0.16} 9}{\color[rgb]{0.8,0.2,0.2} 7}{\color[rgb]{0.8,0.2,0.2} 2}{\color[rgb]{0.31, 0.49, 0.16} 0}{\color[rgb]{0.31, 0.49, 0.16} .}\\
{\bf "Combined" prediction:} & {\color[rgb]{0.31, 0.49, 0.16} -}{\color[rgb]{0.31, 0.49, 0.16} 1}{\color[rgb]{0.31, 0.49, 0.16} 8}{\color[rgb]{0.8,0.2,0.2} 2}{\color[rgb]{0.8,0.2,0.2} 0}{\color[rgb]{0.8,0.2,0.2} 7}{\color[rgb]{0.31, 0.49, 0.16} 0}{\color[rgb]{0.31, 0.49, 0.16} 0}{\color[rgb]{0.31, 0.49, 0.16} .}\\
\end{tabular}

\end{mdframed}\vspace{0.07cm}
\vspace{-0.1cm}\begin{mdframed}\vspace{0.07cm}
{\bf Input:}
\begin{lstlisting}
print((678740+203140)).
\end{lstlisting}
\begin{tabular}{ l l }
{\bf Target:} & 881880. \\
{\bf "Baseline" prediction:} & {\color[rgb]{0.31, 0.49, 0.16} 8}{\color[rgb]{0.31, 0.49, 0.16} 8}{\color[rgb]{0.8,0.2,0.2} 0}{\color[rgb]{0.8,0.2,0.2} 4}{\color[rgb]{0.8,0.2,0.2} 7}{\color[rgb]{0.8,0.2,0.2} 5}{\color[rgb]{0.31, 0.49, 0.16} .}\\
{\bf "Naive" prediction:} & {\color[rgb]{0.31, 0.49, 0.16} 8}{\color[rgb]{0.31, 0.49, 0.16} 8}{\color[rgb]{0.31, 0.49, 0.16} 1}{\color[rgb]{0.8,0.2,0.2} 6}{\color[rgb]{0.8,0.2,0.2} 6}{\color[rgb]{0.8,0.2,0.2} 6}{\color[rgb]{0.31, 0.49, 0.16} .}\\
{\bf "Mix" prediction:} & {\color[rgb]{0.31, 0.49, 0.16} 8}{\color[rgb]{0.31, 0.49, 0.16} 8}{\color[rgb]{0.8,0.2,0.2} 0}{\color[rgb]{0.8,0.2,0.2} 1}{\color[rgb]{0.8,0.2,0.2} 9}{\color[rgb]{0.31, 0.49, 0.16} 0}{\color[rgb]{0.31, 0.49, 0.16} .}\\
{\bf "Combined" prediction:} & {\color[rgb]{0.31, 0.49, 0.16} 8}{\color[rgb]{0.31, 0.49, 0.16} 8}{\color[rgb]{0.8,0.2,0.2} 5}{\color[rgb]{0.8,0.2,0.2} 9}{\color[rgb]{0.8,0.2,0.2} 2}{\color[rgb]{0.31, 0.49, 0.16} 0}{\color[rgb]{0.31, 0.49, 0.16} .}\\
\end{tabular}

\end{mdframed}\vspace{0.07cm}
\vspace{-0.1cm}\begin{mdframed}\vspace{0.07cm}
{\bf Input:}
\begin{lstlisting}
print((929067-75246)).
\end{lstlisting}
\begin{tabular}{ l l }
{\bf Target:} & 853821. \\
{\bf "Baseline" prediction:} & {\color[rgb]{0.31, 0.49, 0.16} 8}{\color[rgb]{0.31, 0.49, 0.16} 5}{\color[rgb]{0.8,0.2,0.2} 1}{\color[rgb]{0.8,0.2,0.2} 2}{\color[rgb]{0.8,0.2,0.2} 3}{\color[rgb]{0.8,0.2,0.2} 3}{\color[rgb]{0.31, 0.49, 0.16} .}\\
{\bf "Naive" prediction:} & {\color[rgb]{0.31, 0.49, 0.16} 8}{\color[rgb]{0.8,0.2,0.2} 6}{\color[rgb]{0.8,0.2,0.2} 7}{\color[rgb]{0.8,0.2,0.2} 1}{\color[rgb]{0.8,0.2,0.2} 1}{\color[rgb]{0.8,0.2,0.2} 3}{\color[rgb]{0.31, 0.49, 0.16} .}\\
{\bf "Mix" prediction:} & {\color[rgb]{0.31, 0.49, 0.16} 8}{\color[rgb]{0.31, 0.49, 0.16} 5}{\color[rgb]{0.8,0.2,0.2} 5}{\color[rgb]{0.8,0.2,0.2} 6}{\color[rgb]{0.8,0.2,0.2} 1}{\color[rgb]{0.8,0.2,0.2} 5}{\color[rgb]{0.31, 0.49, 0.16} .}\\
{\bf "Combined" prediction:} & {\color[rgb]{0.31, 0.49, 0.16} 8}{\color[rgb]{0.31, 0.49, 0.16} 5}{\color[rgb]{0.31, 0.49, 0.16} 3}{\color[rgb]{0.8,0.2,0.2} 0}{\color[rgb]{0.8,0.2,0.2} 0}{\color[rgb]{0.8,0.2,0.2} 9}{\color[rgb]{0.31, 0.49, 0.16} .}\\
\end{tabular}

\end{mdframed}\vspace{0.07cm}
\vspace{-0.1cm}\begin{mdframed}\vspace{0.07cm}
{\bf Input:}
\begin{lstlisting}
d=960350
for x in range(24):d-=187946
print(d).
\end{lstlisting}
\begin{tabular}{ l l }
{\bf Target:} & -3550354. \\
{\bf "Baseline" prediction:} & {\color[rgb]{0.31, 0.49, 0.16} -}{\color[rgb]{0.31, 0.49, 0.16} 3}{\color[rgb]{0.31, 0.49, 0.16} 5}{\color[rgb]{0.8,0.2,0.2} 7}{\color[rgb]{0.8,0.2,0.2} 1}{\color[rgb]{0.8,0.2,0.2} 9}{\color[rgb]{0.8,0.2,0.2} 9}{\color[rgb]{0.8,0.2,0.2} 8}{\color[rgb]{0.31, 0.49, 0.16} .}\\
{\bf "Naive" prediction:} & {\color[rgb]{0.31, 0.49, 0.16} -}{\color[rgb]{0.31, 0.49, 0.16} 3}{\color[rgb]{0.8,0.2,0.2} 6}{\color[rgb]{0.8,0.2,0.2} 9}{\color[rgb]{0.8,0.2,0.2} 9}{\color[rgb]{0.8,0.2,0.2} 9}{\color[rgb]{0.8,0.2,0.2} 9}{\color[rgb]{0.8,0.2,0.2} 3}{\color[rgb]{0.31, 0.49, 0.16} .}\\
{\bf "Mix" prediction:} & {\color[rgb]{0.31, 0.49, 0.16} -}{\color[rgb]{0.31, 0.49, 0.16} 3}{\color[rgb]{0.8,0.2,0.2} 8}{\color[rgb]{0.8,0.2,0.2} 9}{\color[rgb]{0.8,0.2,0.2} 9}{\color[rgb]{0.8,0.2,0.2} 2}{\color[rgb]{0.8,0.2,0.2} 2}{\color[rgb]{0.8,0.2,0.2} 0}{\color[rgb]{0.31, 0.49, 0.16} .}\\
{\bf "Combined" prediction:} & {\color[rgb]{0.31, 0.49, 0.16} -}{\color[rgb]{0.31, 0.49, 0.16} 3}{\color[rgb]{0.31, 0.49, 0.16} 5}{\color[rgb]{0.8,0.2,0.2} 0}{\color[rgb]{0.8,0.2,0.2} 7}{\color[rgb]{0.8,0.2,0.2} 7}{\color[rgb]{0.8,0.2,0.2} 9}{\color[rgb]{0.8,0.2,0.2} 0}{\color[rgb]{0.31, 0.49, 0.16} .}\\
\end{tabular}

\end{mdframed}\vspace{0.07cm}
\vspace{-0.1cm}\begin{mdframed}\vspace{0.07cm}
{\bf Input:}
\begin{lstlisting}
print((8*786463)).
\end{lstlisting}
\begin{tabular}{ l l }
{\bf Target:} & 6291704. \\
{\bf "Baseline" prediction:} & {\color[rgb]{0.31, 0.49, 0.16} 6}{\color[rgb]{0.31, 0.49, 0.16} 2}{\color[rgb]{0.8,0.2,0.2} 7}{\color[rgb]{0.8,0.2,0.2} 0}{\color[rgb]{0.8,0.2,0.2} 8}{\color[rgb]{0.31, 0.49, 0.16} 0}{\color[rgb]{0.31, 0.49, 0.16} 4}{\color[rgb]{0.31, 0.49, 0.16} .}\\
{\bf "Naive" prediction:} & {\color[rgb]{0.31, 0.49, 0.16} 6}{\color[rgb]{0.31, 0.49, 0.16} 2}{\color[rgb]{0.8,0.2,0.2} 7}{\color[rgb]{0.31, 0.49, 0.16} 1}{\color[rgb]{0.8,0.2,0.2} 9}{\color[rgb]{0.31, 0.49, 0.16} 0}{\color[rgb]{0.31, 0.49, 0.16} 4}{\color[rgb]{0.31, 0.49, 0.16} .}\\
{\bf "Mix" prediction:} & {\color[rgb]{0.31, 0.49, 0.16} 6}{\color[rgb]{0.31, 0.49, 0.16} 2}{\color[rgb]{0.31, 0.49, 0.16} 9}{\color[rgb]{0.8,0.2,0.2} 7}{\color[rgb]{0.8,0.2,0.2} 6}{\color[rgb]{0.8,0.2,0.2} 4}{\color[rgb]{0.31, 0.49, 0.16} 4}{\color[rgb]{0.31, 0.49, 0.16} .}\\
{\bf "Combined" prediction:} & {\color[rgb]{0.31, 0.49, 0.16} 6}{\color[rgb]{0.31, 0.49, 0.16} 2}{\color[rgb]{0.8,0.2,0.2} 7}{\color[rgb]{0.8,0.2,0.2} 0}{\color[rgb]{0.8,0.2,0.2} 0}{\color[rgb]{0.31, 0.49, 0.16} 0}{\color[rgb]{0.31, 0.49, 0.16} 4}{\color[rgb]{0.31, 0.49, 0.16} .}\\
\end{tabular}

\end{mdframed}\vspace{0.07cm}
\vspace{-0.1cm}\begin{mdframed}\vspace{0.07cm}
{\bf Input:}
\begin{lstlisting}
print((498592-570324)).
\end{lstlisting}
\begin{tabular}{ l l }
{\bf Target:} & -71732. \\
{\bf "Baseline" prediction:} & {\color[rgb]{0.31, 0.49, 0.16} -}{\color[rgb]{0.8,0.2,0.2} 6}{\color[rgb]{0.31, 0.49, 0.16} 1}{\color[rgb]{0.8,0.2,0.2} 0}{\color[rgb]{0.8,0.2,0.2} 8}{\color[rgb]{0.8,0.2,0.2} 6}{\color[rgb]{0.31, 0.49, 0.16} .}\\
{\bf "Naive" prediction:} & {\color[rgb]{0.31, 0.49, 0.16} -}{\color[rgb]{0.31, 0.49, 0.16} 7}{\color[rgb]{0.8,0.2,0.2} 3}{\color[rgb]{0.8,0.2,0.2} 5}{\color[rgb]{0.8,0.2,0.2} 8}{\color[rgb]{0.31, 0.49, 0.16} 2}{\color[rgb]{0.31, 0.49, 0.16} .}\\
{\bf "Mix" prediction:} & {\color[rgb]{0.31, 0.49, 0.16} -}{\color[rgb]{0.8,0.2,0.2} 1}{\color[rgb]{0.8,0.2,0.2} 9}{\color[rgb]{0.8,0.2,0.2} 0}{\color[rgb]{0.8,0.2,0.2} 0}{\color[rgb]{0.8,0.2,0.2} 0}{\color[rgb]{0.31, 0.49, 0.16} .}\\
{\bf "Combined" prediction:} & {\color[rgb]{0.31, 0.49, 0.16} -}{\color[rgb]{0.31, 0.49, 0.16} 7}{\color[rgb]{0.8,0.2,0.2} 2}{\color[rgb]{0.8,0.2,0.2} 8}{\color[rgb]{0.8,0.2,0.2} 4}{\color[rgb]{0.31, 0.49, 0.16} 2}{\color[rgb]{0.31, 0.49, 0.16} .}\\
\end{tabular}

\end{mdframed}\vspace{0.07cm}
\vspace{-0.1cm}

\vspace{0.3cm}\subsection{Examples of program evaluation prediction. Length = 6, Nesting = 2}\vspace{0.3cm}

\begin{mdframed}\vspace{0.07cm}
{\bf Input:}
\begin{lstlisting}
print((39007+416968)).
\end{lstlisting}
\begin{tabular}{ l l }
{\bf Target:} & 455975. \\
{\bf "Baseline" prediction:} & {\color[rgb]{0.8,0.2,0.2} 5}{\color[rgb]{0.31, 0.49, 0.16} 5}{\color[rgb]{0.8,0.2,0.2} 9}{\color[rgb]{0.31, 0.49, 0.16} 9}{\color[rgb]{0.8,0.2,0.2} 1}{\color[rgb]{0.8,0.2,0.2} 7}{\color[rgb]{0.31, 0.49, 0.16} .}\\
{\bf "Naive" prediction:} & {\color[rgb]{0.31, 0.49, 0.16} 4}{\color[rgb]{0.8,0.2,0.2} 3}{\color[rgb]{0.8,0.2,0.2} 8}{\color[rgb]{0.8,0.2,0.2} 8}{\color[rgb]{0.8,0.2,0.2} 8}{\color[rgb]{0.8,0.2,0.2} 7}{\color[rgb]{0.31, 0.49, 0.16} .}\\
{\bf "Mix" prediction:} & {\color[rgb]{0.31, 0.49, 0.16} 4}{\color[rgb]{0.31, 0.49, 0.16} 5}{\color[rgb]{0.8,0.2,0.2} 8}{\color[rgb]{0.31, 0.49, 0.16} 9}{\color[rgb]{0.8,0.2,0.2} 9}{\color[rgb]{0.8,0.2,0.2} 3}{\color[rgb]{0.31, 0.49, 0.16} .}\\
{\bf "Combined" prediction:} & {\color[rgb]{0.31, 0.49, 0.16} 4}{\color[rgb]{0.31, 0.49, 0.16} 5}{\color[rgb]{0.8,0.2,0.2} 0}{\color[rgb]{0.8,0.2,0.2} 0}{\color[rgb]{0.8,0.2,0.2} 3}{\color[rgb]{0.8,0.2,0.2} 1}{\color[rgb]{0.31, 0.49, 0.16} .}\\
\end{tabular}

\end{mdframed}\vspace{0.07cm}
\vspace{-0.1cm}\begin{mdframed}\vspace{0.07cm}
{\bf Input:}
\begin{lstlisting}
print((586051+664462)).
\end{lstlisting}
\begin{tabular}{ l l }
{\bf Target:} & 1250513. \\
{\bf "Baseline" prediction:} & {\color[rgb]{0.31, 0.49, 0.16} 1}{\color[rgb]{0.31, 0.49, 0.16} 2}{\color[rgb]{0.31, 0.49, 0.16} 5}{\color[rgb]{0.31, 0.49, 0.16} 0}{\color[rgb]{0.8,0.2,0.2} 9}{\color[rgb]{0.8,0.2,0.2} 3}{\color[rgb]{0.8,0.2,0.2} 9}{\color[rgb]{0.31, 0.49, 0.16} .}\\
{\bf "Naive" prediction:} & {\color[rgb]{0.31, 0.49, 0.16} 1}{\color[rgb]{0.31, 0.49, 0.16} 2}{\color[rgb]{0.8,0.2,0.2} 4}{\color[rgb]{0.31, 0.49, 0.16} 0}{\color[rgb]{0.8,0.2,0.2} 7}{\color[rgb]{0.31, 0.49, 0.16} 1}{\color[rgb]{0.8,0.2,0.2} 9}{\color[rgb]{0.31, 0.49, 0.16} .}\\
{\bf "Mix" prediction:} & {\color[rgb]{0.31, 0.49, 0.16} 1}{\color[rgb]{0.31, 0.49, 0.16} 2}{\color[rgb]{0.8,0.2,0.2} 3}{\color[rgb]{0.31, 0.49, 0.16} 0}{\color[rgb]{0.8,0.2,0.2} 8}{\color[rgb]{0.8,0.2,0.2} 8}{\color[rgb]{0.8,0.2,0.2} 1}{\color[rgb]{0.31, 0.49, 0.16} .}\\
{\bf "Combined" prediction:} & {\color[rgb]{0.31, 0.49, 0.16} 1}{\color[rgb]{0.31, 0.49, 0.16} 2}{\color[rgb]{0.8,0.2,0.2} 4}{\color[rgb]{0.31, 0.49, 0.16} 0}{\color[rgb]{0.31, 0.49, 0.16} 5}{\color[rgb]{0.8,0.2,0.2} 5}{\color[rgb]{0.8,0.2,0.2} 1}{\color[rgb]{0.31, 0.49, 0.16} .}\\
\end{tabular}

\end{mdframed}\vspace{0.07cm}
\vspace{-0.1cm}\begin{mdframed}\vspace{0.07cm}
{\bf Input:}
\begin{lstlisting}
print(948950).
\end{lstlisting}
\begin{tabular}{ l l }
{\bf Target:} & 948950. \\
{\bf "Baseline" prediction:} & {\color[rgb]{0.31, 0.49, 0.16} 9}{\color[rgb]{0.31, 0.49, 0.16} 4}{\color[rgb]{0.31, 0.49, 0.16} 8}{\color[rgb]{0.31, 0.49, 0.16} 9}{\color[rgb]{0.31, 0.49, 0.16} 5}{\color[rgb]{0.31, 0.49, 0.16} 0}{\color[rgb]{0.31, 0.49, 0.16} .}\\
{\bf "Naive" prediction:} & {\color[rgb]{0.31, 0.49, 0.16} 9}{\color[rgb]{0.31, 0.49, 0.16} 4}{\color[rgb]{0.31, 0.49, 0.16} 8}{\color[rgb]{0.31, 0.49, 0.16} 9}{\color[rgb]{0.31, 0.49, 0.16} 5}{\color[rgb]{0.31, 0.49, 0.16} 0}{\color[rgb]{0.31, 0.49, 0.16} .}\\
{\bf "Mix" prediction:} & {\color[rgb]{0.31, 0.49, 0.16} 9}{\color[rgb]{0.31, 0.49, 0.16} 4}{\color[rgb]{0.31, 0.49, 0.16} 8}{\color[rgb]{0.31, 0.49, 0.16} 9}{\color[rgb]{0.31, 0.49, 0.16} 5}{\color[rgb]{0.31, 0.49, 0.16} 0}{\color[rgb]{0.31, 0.49, 0.16} .}\\
{\bf "Combined" prediction:} & {\color[rgb]{0.31, 0.49, 0.16} 9}{\color[rgb]{0.31, 0.49, 0.16} 4}{\color[rgb]{0.31, 0.49, 0.16} 8}{\color[rgb]{0.31, 0.49, 0.16} 9}{\color[rgb]{0.31, 0.49, 0.16} 5}{\color[rgb]{0.31, 0.49, 0.16} 0}{\color[rgb]{0.31, 0.49, 0.16} .}\\
\end{tabular}

\end{mdframed}\vspace{0.07cm}
\vspace{-0.1cm}\begin{mdframed}\vspace{0.07cm}
{\bf Input:}
\begin{lstlisting}
i=849846
for x in range(15):i-=557574
print((362961 if 881013<597832 else i)).
\end{lstlisting}
\begin{tabular}{ l l }
{\bf Target:} & -7513764. \\
{\bf "Baseline" prediction:} & {\color[rgb]{0.31, 0.49, 0.16} -}{\color[rgb]{0.31, 0.49, 0.16} 7}{\color[rgb]{0.8,0.2,0.2} 4}{\color[rgb]{0.8,0.2,0.2} 2}{\color[rgb]{0.8,0.2,0.2} 2}{\color[rgb]{0.31, 0.49, 0.16} 7}{\color[rgb]{0.8,0.2,0.2} 5}{\color[rgb]{0.8,0.2,0.2} 6}{\color[rgb]{0.31, 0.49, 0.16} .}\\
{\bf "Naive" prediction:} & {\color[rgb]{0.31, 0.49, 0.16} -}{\color[rgb]{0.31, 0.49, 0.16} 7}{\color[rgb]{0.8,0.2,0.2} 0}{\color[rgb]{0.31, 0.49, 0.16} 1}{\color[rgb]{0.8,0.2,0.2} 1}{\color[rgb]{0.8,0.2,0.2} 0}{\color[rgb]{0.8,0.2,0.2} 4}{\color[rgb]{0.8,0.2,0.2} 8}{\color[rgb]{0.31, 0.49, 0.16} .}\\
{\bf "Mix" prediction:} & {\color[rgb]{0.31, 0.49, 0.16} -}{\color[rgb]{0.8,0.2,0.2} 2}{\color[rgb]{0.8,0.2,0.2} 6}{\color[rgb]{0.31, 0.49, 0.16} 1}{\color[rgb]{0.8,0.2,0.2} 7}{\color[rgb]{0.31, 0.49, 0.16} 7}{\color[rgb]{0.8,0.2,0.2} 7}{\color[rgb]{0.8,0.2,0.2} 7}{\color[rgb]{0.31, 0.49, 0.16} .}\\
{\bf "Combined" prediction:} & {\color[rgb]{0.31, 0.49, 0.16} -}{\color[rgb]{0.31, 0.49, 0.16} 7}{\color[rgb]{0.8,0.2,0.2} 1}{\color[rgb]{0.8,0.2,0.2} 0}{\color[rgb]{0.8,0.2,0.2} 1}{\color[rgb]{0.8,0.2,0.2} 1}{\color[rgb]{0.8,0.2,0.2} 4}{\color[rgb]{0.8,0.2,0.2} 6}{\color[rgb]{0.31, 0.49, 0.16} .}\\
\end{tabular}%

\end{mdframed}\vspace{0.07cm}
\vspace{-0.1cm}

\vspace{0.3cm}\subsection{Examples of program evaluation prediction. Length = 6, Nesting = 3}\vspace{0.3cm}

\begin{mdframed}\vspace{0.07cm}
{\bf Input:}
\begin{lstlisting}
c=335973;
b=(c+756088);
print((6*(b+66858))).
\end{lstlisting}
%

\end{mdframed}\vspace{0.07cm}
\vspace{-0.1cm}

\vspace{0.3cm}\subsection{Examples of predicting result of addition.\\Length = 6}\vspace{0.3cm}

\begin{mdframed}\vspace{0.07cm}
{\bf Input:}
\begin{lstlisting}
print(284993+281178).
\end{lstlisting}
%

\end{mdframed}\vspace{0.07cm}
\vspace{-0.1cm}

\vspace{0.3cm}\subsection{Examples of predicting result of addition.\\Length = 8}\vspace{0.3cm}
\begin{mdframed}\vspace{0.07cm}
{\bf Input:}
\begin{lstlisting}
print(32847917+95908452).
\end{lstlisting}
%

\end{mdframed}\vspace{0.07cm}
\vspace{-0.1cm}

\end{document}